\pdfoutput=1

\documentclass[11pt]{article}

\usepackage[]{acl}

\usepackage{times}
\usepackage{latexsym}
\usepackage{natbib}
\usepackage{subcaption}

\usepackage{float}
\usepackage{ulem}

\usepackage[T1]{fontenc}

\usepackage[utf8]{inputenc}

\usepackage{microtype}

\usepackage{inconsolata}

\usepackage{booktabs}       
\usepackage{amsfonts}       
\usepackage{nicefrac}       
\usepackage{xcolor}         
\usepackage{enumitem}
\usepackage{multicol}
\usepackage{multirow}
\usepackage{array}
\usepackage{graphicx}
\usepackage{wrapfig}
\usepackage{latexsym}
\usepackage{amsmath}
\usepackage{listings}
\usepackage{mdframed}
\mdfsetup{nobreak=true}
\usepackage[dvipsnames]{xcolor}
\usepackage{titletoc}

\newcommand\ilsiclay{\texttt{ILSIC-Lay}}
\newcommand\ilsicmulti{\texttt{ILSIC-Multi}}
\newcommand{\stsuf}[1]{{#1}$_{stat}$}
\newcommand{\trsuf}[1]{{#1}$_{train}$}
\newcommand{\vdsuf}[1]{{#1}$_{val}$}
\newcommand{\tesuf}[1]{{#1}$_{test}$}
\newcommand{\ltrsuf}[1]{{#1}$_{train}^{lay}$}
\newcommand{\lvdsuf}[1]{{#1}$_{val}^{lay}$}
\newcommand{\ctrsuf}[1]{{#1}$_{train}^{court}$}
\newcommand{\cvdsuf}[1]{{#1}$_{val}^{court}$}

\definecolor{gpt-color}{HTML}{D10000}
\definecolor{llama-color}{HTML}{1716C7}
\definecolor{gemma-color}{HTML}{198A2C}

\title{ILSIC: Corpora for Identifying Indian Legal Statutes\\ from Queries by Laypeople}

\author{{\bf Shounak Paul}
\thanks{\ \ Equal Contributions} \qquad
{\bf Raghav Dogra}\footnotemark[1] 
\qquad  
\textbf{Pawan Goyal} \qquad \textbf{Saptarshi Ghosh} \\  
        Indian Institute of Technology, Kharagpur\\
  \texttt{\{shounakpaul95,raghav25t\}@kgpian.iitkgp.ac.in}, \\
  \texttt{\{pawang,saptarshi\}@cse.iitkgp.ac.in} \\  
}

\begin{document}
\maketitle

\begin{abstract}
Legal Statute Identification (LSI) for a given situation is one of the most fundamental tasks in Legal NLP. This task has traditionally been modeled using facts from court judgments as input queries, due to their abundance. However, in practical settings, the input queries are likely to be informal and asked by laypersons, or non-professionals. 
While a few laypeople LSI datasets exist, there has been little research to explore the differences between court and laypeople data for LSI.
In this work, we create \textbf{\texttt{ILSIC}}, a corpus of laypeople queries covering 500+ statutes from Indian law. Additionally, the corpus also contains court case judgements to enable researchers to effectively compare between court and laypeople data for LSI. We conducted extensive experiments on our corpus, including benchmarking over the laypeople dataset using zero and few-shot inference, retrieval-augmented generation and supervised fine-tuning. We observe that models trained purely on court judgements are ineffective during test on laypeople queries, while transfer learning from court to laypeople data can be beneficial in certain scenarios. We also conducted fine-grained analyses of our results in terms of categories of queries and frequency of statutes. 
\end{abstract}
\section{Introduction} \label{sec:introduction}


\noindent In the legal domain, laws (statutes) are considered to be one of the most fundamental sources of knowledge that guide principles of jurisdiction~\cite{joshi2023ucreat}. 
In practice, a legal practitioner when faced with a legal situation, typically uses their experience and knowledge to identify the applicable statutes, which can be a time-consuming activity. 
Hence, the task of automated Legal Statute Identification (LSI) has gained prominence.

Early LSI methods relied on statistical or shallow ML models over small datasets~\citep{salton1988termweight,zeng2007knowledge}, but deep learning spurred the creation of large-scale datasets from court judgments, where case facts serve as queries and cited statutes as gold-standard outputs~\citep{xiao2018cail2018,chalkidis2021echr,paul2022ilsi,chalkidis2019neural,zhong2018topjudge,wang2019hmn}. Other sources, like bar exam questions, have also been used~\citep{li2023coliee23,li2024coliee2024}. These, however, differ linguistically from real laypeople queries, which are informal, ambiguous, and lack legal terminology~\citep{su2024stard}.

Recent work addresses this gap by building datasets from queries by citizens or laypeople - STARD for Chinese~\citep{su2024stard}, BSARD/BiBSARD for Belgian law~\citep{louis2021bsard,lotfi2024bibsard}, and ALQAC for Vietnamese law~\citep{nguyen2023alqac23,do2024alqac24}. Yet, no study has directly compared judgment-based vs. laypeople-query datasets for LSI; \citet{su2024stard} only reported limited indirect gains by using laypeople data with LLMs in other downstream tasks.

To address this research gap, we make the following contributions in this resource paper:
\begin{itemize}[noitemsep,nosep,leftmargin=*]

\item  We develop \textbf{\texttt{ILSIC}} (Indian Legal Statute Identification Corpus), a novel collection of datasets based on both laypeople queries as well as court judgment facts for Indian law. 
Our corpus consists of two independent datasets --

\noindent (i)~\textbf{\ilsiclay{}:} A dataset containing 8K laypeople queries covering 567 Indian statutes, meant to serve as the first benchmark for laypeople LSI in the Indian domain;

\noindent (ii)~\textbf{\ilsicmulti{}:} A dataset containing two different groups of train/validation sets (one based on court facts, the other based on laypeople queries) and a single test set of laypeople queries, all covering a common set of 399 Indian statutes. This corpus allows researchers to directly compare between models being trained on laypeople vs. court data, or any combination of these.

\item We run benchmarking experiments on \ilsiclay{} with traditional retrieval methods, such as BM25~\citep{robertson2009bm25}, SentenceBERT (SBERT)~\citep{reimers2019sentencebert} and SAILER~\citep{li2023sailer}, as well as state-of-the-art Large Language Models (LLMs) such as GPT-4.1, Llama-3 and Gemma-3, using various experimental setups/variations such as zero and few-shot prompting, retrieval-augmented generation (RAG) and supervised fine-tuning (SFT). Our results demonstrate the relative difficulty of the LSI task, with either GPT-4.1 or fine-tuned versions of the other LLMs not crossing 35\% $\mu$F1.

\item We utilize the \ilsicmulti{} corpus to compare between court facts vs. laypeople queries as input. We devise multiple settings, such as Zero-Shot Inference and Supervised Fine-tuning on either court or laypeople data. We observe that pure court-tuning is completely inefficient for answering laypeople queries. 
We also created some synthetic laypeople queries by processing court facts with an LLM; this setting can be thought to be somewhere between pure court facts and real-life laypeople queries. However, there is no improvement over pure court data in this setting. We also conducted a transfer learning experiment by sequentially training on court data first followed by laypeople data, which showed limited improvement for only Llama-3.

\item We conduct extensive analyses of the above results, by analyzing the model performances across different frequencies of statutes and query categories. We observe that all models have reduced performance over the rare statutes, while the correlation between F1-score and query category frequency is not as strong. 
\end{itemize}
All the datasets, codes and trained model parameters are available at \url{https://github.com/Law-AI/ilsic}.

\section{Related Work} \label{sec:relwork}

\begin{table*}[ht]
\centering
\small
\addtolength{\tabcolsep}{-0.1em}
\begin{tabular}{@{}p{4.2cm} p{1.5cm} p{4.5cm} p{1.2cm} p{3cm}@{}}
\toprule
\textbf{Dataset} & \textbf{Jurisdiction} & \textbf{Input Type} & \textbf{\#Queries} & \textbf{\#Statutes} \\ \midrule
\multicolumn{5}{c}{\bf Prior Datasets using formal queries} \\ \midrule
ILSI~\citep{paul2022ilsi} & India & Case-facts from judgments & 66k & 100 IPC statutes \\
CAIL2018~\citep{xiao2018cail2018} & China & Criminal case judgments & 2.68M & 183 Chinese criminal articles \\
COLIEE Task 3~\citep{li2023coliee23,li2024coliee2024} & Japan & Bar exam-style queries & 996 train / 101 test & 768 Japanese laws \\
ECHR~\citep{chalkidis2021echr} & EU & Case-facts from ECHR judgments & 11k & 10 ECHR statutes \\ \midrule
\multicolumn{5}{c}{\bf Prior Datasets using informal/laypeople queries} \\ \midrule
STARD~\citep{su2024stard} & China & Laypeople queries from legal consultations & 1543 & 55k Chinese articles \\
BSARD~\citep{louis2021bsard} & Belgium & Legal queries by citizens & 1100 & 22.6k Belgian laws \\
ALQAC~\citep{nguyen2023alqac23,do2024alqac24} & Vietnam & Synthetic legal scenarios crafted by laypeople  & 9846 & Various \\ \midrule
\multicolumn{5}{c}{\bf \texttt{ILSIC} (This work)} \\ \midrule
\ilsiclay{} & India & Legal queries by laypeople & 8k & 569 Indian statutes \\
\ilsicmulti{} & India & Legal queries by laypeople + case-facts from Indian judgments & 7k & 399 Indian statutes \\
\bottomrule
\end{tabular}
\caption{A brief comparison of our \texttt{ILSIC} datasets with prior works}
\end{table*}

\noindent \underline{\textbf{LSI using formal legal documents:}}
Traditionally, court case judgments have been used for creating LSI datasets for countries that follow some degree of the civil law system. Such case judgments contain the facts, i.e., the necessary background information which serves as the input, as well as references to the statutes considered relevant to the case by legal professionals which serves as the gold-standard output.
\citet{xiao2018cail2018} introduced the CAIL2018, the first large-scale dataset for Chinese criminal law, comprising of over 2.6M samples and 183 statutes.
\citet{chalkidis2019neural} introduced the ECHR dataset based on case facts from the European Court of Human Rights, which involved a binary prediction of whether any statute has been violated. A newer version of the dataset involved multi-label classification over 10 ECHR statutes~\citep{chalkidis2021echr}.
\citet{paul2022ilsi} introduced the ILSI dataset for identifying 100 criminal statutes from the Indian Penal Code.
While these datasets were created using court judgment documents, the COLIEE Task 3 family of datasets~\citep{li2023coliee23,li2024coliee2024} require retrieving the relevant statutes from Japanese law given questions from the Bar exam.

\vspace{2mm}
\noindent \underline{\textbf{LSI using informal queries:}} Although using formal legal texts such as judgment documents can be used as a proxy for modeling LSI, in the practical setting, the automated method should be able to process queries/facts written in an informal way, by laypeople or non-professionals. Lately, some works have focused on creating LSI datasets from such laypeople queries.
\citet{su2024stard} released STARD, a Chinese LSI dataset based on real-world legal consultation data.
\citet{louis2021bsard} introduced BSARD, a French LSI dataset based on Belgian law, consisting of legal queries by Belgian citizens. 
This work was later extended to a bilingual setting, through the addition of queries in Dutch~\citep{lotfi2024bibsard}.
The ALQAC Task 1 family of datasets~\citep{nguyen2023alqac23,do2024alqac24} require retrieving the relevant statutes from Vietnamese law, given description of legal scenarios in Vietnamese and Thai language.
For the Indian legal scenario, \citet{bhattacharya2019aila} introduced a very limited set of 50 real-world legal scenarios, annotated with statutes from the Indian Penal Code.
Among all these studies, only \citet{su2024stard} have made a comparison between formal and informal queries, demonstrating that utilizing laypeople data as the knowledge source of RAG can improve performance of LLMs on other downstream tasks. However, no study, to the best of our knowledge, has conducted a direct comparison between using laypeople queries vs. court queries for the task of LSI.

\section{Corpora for Legal Statute Identification} \label{sec:corpus}

As discussed in Section~\ref{sec:introduction}, the objective of this study is two-fold:
(i)~create the first LSR dataset for Indian law that are based on real-life queries asked by laypeople (non-professionals), and 
(ii)~compare the effects of using laypeople queries vis-a-vis queries created from more authentic legal sources such as court judgment documents.
To this end, we create two separate, but related corpora:

\vspace{1mm}
\noindent (i)~\ilsiclay{}: Consists of a total of 8127 queries ranging across a set of 569 statutes \stsuf{\ilsiclay}. All the queries in train, validation and test splits are sourced from real-life queries asked by laypersons on an online legal forum. 
This is the general dataset for performing LSI with laypeople queries, and consists of a range of statutes, including very rare ones.

\vspace{1mm}
\noindent (ii)~\ilsicmulti{}: Consists of two distinct groups of train and validation sets, and a single test set based on laypeople queries. The entire corpus spans a set of 399 statutes \stsuf{\ilsicmulti}, which is a subset of \stsuf{\ilsiclay}. One group of train and validation sets, namely \ltrsuf{\ilsicmulti} and \lvdsuf{\ilsicmulti} are subsets of \trsuf{\ilsiclay} and \vdsuf{\ilsiclay} respectively, and are to be used for training models with laypeople queries. The other group of train and validation sets, namely \ctrsuf{\ilsicmulti} and \cvdsuf{\ilsicmulti}, are based on facts extracted from court judgment documents. The sole test set \tesuf{\ilsicmulti} consists of laypeople queries and is a subset of \tesuf{\ilsiclay}. The purpose of this suite of datasets is to evaluate the effectiveness of using laypeople vs. court queries to train models over the same set of statutes, when the end-task is to identify statutes from real-life laypeople queries.

A brief snapshot of our corpus statistics are presented in Table~\ref{tab:ilsic-stats}.

\subsection{Construction of \ilsiclay} \label{sec:corpus/laypeople}

The Indian law forum \url{https://kaanoon.com} provides facilities for common users (laypersons) to post any legal query, which is then answered by different lawyers associate with the platform.
Each such conversation occurs in the form of a thread, with the user posting their query and lawyers replying to them. 
The user can ask follow-up queries as well.
We scraped $\sim$ 30k such conversations from the website to build the laypeople query datasets.

\begin{table}
    \centering
    \small
    \begin{tabular}{lccc} \toprule
 \textbf{Data Split}& \textbf{\#Queries}&\textbf{\#Words} & \textbf{\#Citations}\\\toprule
         \multicolumn{4}{c}{\textbf{\ilsiclay}} \\\midrule
         \trsuf{\ilsiclay}&  6465& 180.71&2.20\\
         \vdsuf{\ilsiclay}&  826& 184.69&2.32\\
         \tesuf{\ilsiclay}&  836& 180.78&2.40\\ \midrule
         \multicolumn{4}{c}{\textbf{\ilsicmulti}} \\ \midrule
         \ltrsuf{\ilsicmulti}&  5793&  184.62&2.14\\
         \lvdsuf{\ilsicmulti}&  735&  187.20&2.29\\
         \ctrsuf{\ilsicmulti}&  12930& 711.47&2.39\\          \cvdsuf{\ilsicmulti}&  1652& 747.38&2.59\\
         \tesuf{\ilsicmulti}&  757& 183.10&2.33\\ \bottomrule
    \end{tabular}
    \caption{Brief Statistics of \texttt{ILSIC} data splits}
    \label{tab:ilsic-stats}
    
\end{table}

\vspace{1mm}
\noindent \underline{\textbf{Statute Extraction and Normalization:}}
Our source of the gold-standard statutes for each asked query comes from the different lawyer responses.
Some of the user queries themselves may mention a statute, such as, ``I got charged under IPC 279 after crashing my car and injuring some pedestrians \ldots'', and we leave filter those out using regular expressions.
For the rest of the queries, we developed a prompting pipeline with GPT-3.5 Turbo to extract all statute references from the lawyer answers (Prompt provided in App. Table~\ref{tab:legal-extraction-prompt}).
Since lawyers may reference the same statute in different ways, such as ``Section 185 of the Motor Vehicle Act'' or ``s.185 MV Act'' (and many other such variations), we had to make extensive use of regular expressions and fuzzy string matching to normalize the statute names to the most formal one (in the above case, it would be ``Section 185 of the Motor Vehicles Act, 1988'').
By following the above process, we ended with a total set of 600+ statutes.

\vspace{1mm}
\noindent \underline{\textbf{Query Cleaning and Filtering:}} 
Some queries did not have statute references in the respective lawyer answers, and we filtered out all such queries.
Some queries were not in English (either containing characters from non-English alphabets, or transliterated into English from other languages), and these were skipped too. 
After these steps, we had a total of $\sim$ 12k queries spanning 594 statutes, where each query had atleast one statute citation in the corresponding lawyer answers.
We did a final round of filtration, such that finally each statute had at least three supporting queries.
This gave us the final dataset of 8k queries and 567 statutes.
For each of the queries, we used the OpenNyAI model~\citep{kalamkar2022legalner} to detect and mask named entities, to prevent biases for downstream models.

\vspace{1mm}
\noindent \underline{\textbf{Dataset Quality Validation:}} 
We asked a Law expert (a Master in Law graduate from a reputed Law school in India) to manually identify all statutes mentioned in 50 randomly sampled documents. Then we compared the set of statutes identified by the Law expert and the set of statutes identified using the regex approach. Through this manual verification of 50 documents, we confirmed that the regex approach successfully captured over 95\% of statute references. The few missed ones were due to non-standard abbreviations, typing errors, and so on. This demonstrates that the coverage of our extraction pipeline is dependable. While a small fraction of references may not be captured due to formatting inconsistencies, we believe their impact on overall evaluation remains negligible.

\vspace{1mm}
\noindent \underline{\textbf{Dataset Splits and Statistics:}}
For our train/validation/test splits, we wanted to ensure that some statutes are held-out in the validation and test sets, i.e., some statutes are not cited at all in the train set gold-standard. 
We randomly selected some statutes as held-out, and tried to find combinations of train/validation/test splits such that the intended ratio between the three splits is approximately 8:1:1 respectively. 
After multiple iterations, we found a certain combination that satisfied all our criteria. 
This gave us split sizes of 6465 (train), 826 (validation) and 836 (test), citing a total of 535, 367 and 358 statutes respectively.
8 different statutes are held-out in the validation and test sets each.
The queries comprise a total of 10 different legal topics, such as Business Law, Family Law, Taxation, and so on, as tagged by the source law forum (see Table~\ref{tab:ilsic-laypeople-category} in App.~\ref{app:dataset}).  
Some salient statistics of our dataset are summarized in Table~\ref{tab:ilsic-stats}.

\subsection{Construction of \ilsicmulti} \label{sec:corpus/multi}

To conduct analyses regarding the impact of using laypeople vs. court queries for training models, we need to create training corpora for the two different types of queries that are based on the same candidate statute space.
We firstly scraped $\sim$ 50k case judgments from the Supreme Court of India and several prominent High Courts. 
Thereon, we found out the common statutes between those cited across \ilsiclay{} and these court judgments. 
This gives us \stsuf{\ilsicmulti}, a set of 399 statutes that are common to both.

\vspace{2mm}
\noindent \underline{\textbf{Laypeople Query Components:}}
To create \ltrsuf{\ilsicmulti}, we basically filtered out those queries from \trsuf{\ilsiclay} that cite at least one statute from \stsuf{\ilsicmulti}, giving us a total of 5793 queries. 
We repeat the same process over the validation and test sets of \ilsiclay{} to give us \lvdsuf{\ilsicmulti} and \tesuf{\ilsicmulti}, containing 735 and 757 queries respectively.

\vspace{2mm}
\noindent \underline{\textbf{Court Query Components:}}
For preparing the court-based queries, we wanted to create equivalent train and validation sets as \ltrsuf{\ilsicmulti} and \lvdsuf{\ilsicmulti}, but based on court judgment facts. 
To reflect the real-world scenario where court judgment documents are found in abundance compared to laypeople queries, we sampled 12930 court documents for \ctrsuf{\ilsicmulti}, which is the double the size of \ltrsuf{\ilsicmulti}, and spans the same set of 399 statutes. 
We follow the same procedure for creating \cvdsuf{\ilsicmulti}.
For extracting only the fact parts of these judgments, and performing anonymization, we use the OpenNyAI model~\citep{kalamkar2022legalner}. 
The court queries are significantly longer as compared to laypeople queries, as seen in Table~\ref{tab:ilsic-stats}.
A detailed analysis of the linguistic differences between court and laypeople data is presented in App.~\ref{app:dataset}. 
A clear sample demonstrating the difference between court and laypeople data is presented in App.~\ref{app:laypeople-court-examples}.

\section{Methods and General Experimental Settings} \label{sec:method}
As mentioned in Section~\ref{sec:introduction}, the goal of this study is to benchmark different models on the \ilsiclay{} dataset, and perform comparisons between court and laypeople queries with the \ilsicmulti{} dataset.
In this Section, we describe the general models and settings used for the different kind of experiments.

\subsection{Retrieval-based Methods} \label{sec:method/retrieval}

We used multiple retrieval-based methods for benchmarking. 

\vspace{1mm}
\noindent \textbf{BM25~\citep{robertson2009bm25}:} A sparse information retrieval method based on keyword matching.

\vspace{1mm}
\noindent\textbf{SBERT~\citep{dettmers2023qlora}:} A dense retrieval approach that uses a BERT-based encoder trained on sentence similarity, to encode queries and statute definitions into a shared semantic embedding space, which are then compared using dot product (\url{ https://huggingface.co/sentence-transformers/all-mpnet-base-v2}).

\vspace{1mm}
\noindent \textbf{SAILER~\citep{li2023sailer}:} Another dense retrieval model with a BERT-based encoder that has been pre-trained over legal documents and fine-tuned for legal retrieval (\url{https://huggingface.co/CSHaitao/SAILER_en_finetune}).

\vspace{2mm}
\noindent
\underline{\bf Experimental Setup:} These retrieval methods return, for a given query $q$, a ranked list of all the candidates (statutes) based on semantic similarity with $q$. To compare with other approaches, we consider only the top $k$ statutes for each query to be relevant predictions, and the optimal choice of $k$ is determined based on best performance over the respective validation sets.

\subsection{Generative Methods} \label{sec:method/generative}

We also benchmarked several modern generative Large Language Models (LLMs) on our dataset.

\vspace{1mm}
\noindent \textbf{GPT:} We used OpenAI's GPT-4.1 (\url{https://platform.openai.com/docs/models/gpt-4.1}), one of the latest and most powerful versions of GPT, for inference experiments. This is one of the latest and most powerful GPT models.

\vspace{1mm}
\noindent \textbf{LLaMA~\citep{dubey2024llama3}:} LLaMa is one of the most popular open-source LLMs deployed heavily in recent research. We used Llama-3 (8B) (\url{https://huggingface.co/meta-llama/Meta-Llama-3-8B-Instruct}) under different experimental setups listed below.

\vspace{1mm}
\noindent \textbf{Gemma~\citep{kamath2025gemma3}:} Another popular open-source model, we used Gemma-3 (12B) (\url{https://huggingface.co/google/gemma-3-12b-it}).

\vspace{1mm}
\noindent \textbf{Qwen-3:} We used Qwen-3 (4B) (\url{https://huggingface.co/Qwen/Qwen3-4B}) for inference experiments. Qwen-3 is a recent multilingual LLM with strong instruction-following capabilities.

\vspace{1mm}
\noindent \textbf{Phi-4 Mini:} We used Phi-4 Mini (4B) (\url{https://huggingface.co/microsoft/phi-4-mini-instruct}) which is a compact model designed for efficient instruction-following.

\vspace{2mm}
\noindent
\underline{\bf Experimental Setups:}
We use the LLMs in different experimental setups, which we describe below.

\vspace{1mm}
\noindent
\textbf{Zero-Shot Inference (ZSI):} Models were given the prompt to return the most plausible statutes (Section number with Act name) for each query. Most modern LLMs already know most of the statute definitions, which we manually verified for 30\% of the most rarely cited statutes.

\vspace{1mm}
\noindent
\textbf{Two-Shot Inference (TSI):} For a given query, we selected two similar training queries based on SentenceBERT (SBERT) similarity from the same category. These examples were then provided in the prompt along with their corresponding gold-standard statutes.

\vspace{1mm}
\noindent \textbf{Retrieval-Augmented Generation (RAG):} In this setup, we wished to augment the prompt with statute information of some possibly relevant statute candidates. In the setup used in this paper, only statute titles are provided. To identify the possibly relevant statutes, we first used SBERT to rank the training queries similar to fetching two-shot examples (described above), and collected the gold-standard statutes cited by these queries to get a set of 15 statutes. This approach is an indirect method to obtain possibly relevant statutes, since in Indian law, similar queries are likely to cite similar statutes.

\vspace{1mm}
\noindent
\textbf{Supervised Fine-tuning (SFT):}
We also performed supervised fine-tuning using QLoRA~\citep{dettmers2023qlora} with open-source LLMs.

\vspace{1mm}
\noindent All generative models, during inference, require a verbalizer that can match the LLM output with the set of target statutes in our dataset.
We employ a Verbalization Algorithm for LLM Inference that standardizes free-form outputs into canonical \textit{Section--Act} pairs using hybrid fuzzy matching to ensure consistent and comparable statute references across models. (see App.~\ref{app:verb} for details).

The hyper-parameter settings for the different experiments and the prompts used for the LLMs are listed in App.~\ref{app:hyperparams} and App.~\ref{app:prompts}, Table~\ref{tab:llm-prompt} respectively. 

\subsection{Evaluation Metrics}
We evaluated all the experiments under a standard multi-label classification setting, where each query may be associated with multiple applicable statutes. We report both macro- and micro-averaged Precision, Recall, and F1-score. While macro-averaging gives equal weight to each statute, micro-averaging aggregates prediction counts across statutes, and is thus dominated by the frequent statutes.
  

\section{Experiments on \ilsiclay{}} \label{sec:expt-laypeople}

\begin{table}[h]
    \small
    \centering
    \addtolength{\tabcolsep}{-0.22em}
    \begin{tabular}{l|cc|cc|cc}\toprule
         \multirow{2}{*}{\textbf{Model}}&  \multicolumn{2}{c|}{\textbf{Precision}}&  \multicolumn{2}{c|}{\textbf{Recall}}&  \multicolumn{2}{c}{\textbf{F1}}\\
 & \textit{\bf m}& $\mathbf{\mu}$& \textit{\bf m}& $\mathbf{\mu}$& \textit{\bf m}&$\mathbf{\mu}$\\\midrule
         \multicolumn{7}{c}{\textbf{Retrieval-based Methods}}\\ \midrule
         BM25&  2.81&  2.81&  7.79&  6.05&  3.81& 3.84\\
         SBERT&  8.05 &  8.05&  13.94 &  10.24& 9.19 &  9.02\\
         SAILER&  4.37&  4.37&  13.45&  10.92&  6.03& 6.24\\ \midrule
         \multicolumn{7}{c}{\textbf{Zero-Shot Inference}}\\ \midrule
         GPT-4.1&  13.84&  25.75&  \underline{\textcolor{gpt-color}{\textbf{19.51}}}&  \underline{\textcolor{gpt-color}{\textbf{40.91}}}& 14.41& 31.60\\
 Llama-3& 7.77& 15.17& 9.40& 22.23& 7.04&18.03\\
 Gemma-3& 8.98& 20.77& 8.32& 22.28& 6.94&21.50\\ 
 Qwen-3& 4.88& 8.04& 5.89& 16.73& 3.97&10.86\\ 
 Phi-4 Mini& 5.48& 16.79& 4.94& 11.10& 4.22&13.37\\ \midrule
         \multicolumn{7}{c}{\textbf{Two Shot Inference}}\\ \midrule
         GPT-4.1& \underline{\textcolor{gpt-color}{\textbf{15.52}}} & \textcolor{gpt-color}{\textbf{33.10}}& 17.26& 35.43& \underline{\textcolor{gpt-color}{\textbf{14.83}}}& \underline{\textcolor{gpt-color}{\textbf{34.23}}}\\
         Llama-3& 9.42& 17.96& 11.56& 27.32& 9.05& 21.67\\
         Gemma-3& 10.83& 29.87& 10.31& 29.86& 9.58& 29.87\\ \midrule
 \multicolumn{7}{c}{\textbf{Retrieval Augmented Generation}}\\ \midrule
 GPT-4.1& 11.83 & 28.34 & 16.30& 37.92& 12.53& 32.44\\
 LLaMA3 & 8.56 & 23.06 & 7.83  & 24.84 & 7.30 & 23.92 \\
 Gemma3 & 10.60 & 25.78 & \textcolor{gemma-color}{\textbf{13.54}} & \textcolor{gemma-color}{\textbf{34.57}} & 10.63 & 29.54 \\ \midrule
 \multicolumn{7}{c}{\textbf{Supervised Fine-tuning on \trsuf{\ilsiclay}}}\\ \midrule
 Llama-3& \textcolor{llama-color}{\textbf{12.11}}& \textcolor{llama-color}{\textbf{31.69}}& \textcolor{llama-color}{\textbf{12.41}}& \textcolor{llama-color}{\textbf{35.76}}& \textcolor{llama-color}{\textbf{10.91}}& \textcolor{llama-color}{\textbf{31.52}}\\
 Gemma-3& \textcolor{gemma-color}{\textbf{11.29}}& \underline{\textcolor{gemma-color}{\textbf{35.34}}}& 10.98& 31.28& \textcolor{gemma-color}{\textbf{9.95}}& \textcolor{gemma-color}{\textbf{33.19}}\\ \midrule
 \bottomrule
    \end{tabular}
    \caption{Performance of all methods on the \ilsiclay{} dataset. All results are in percentage, and all models have been evaluated on \tesuf{\ilsiclay}. Best values for \textcolor{gpt-color}{\textbf{GPT}}, \textcolor{llama-color}{\textbf{Llama}} and \textcolor{gemma-color}{\textbf{Gemma}} are marked with the respective colours, and the best values overall are underlined.}
    \label{tab:laypeople-results}
\end{table}

We benchmark all the models described above on \ilsiclay{} and report the results in Table~\ref{tab:laypeople-results}. 

\subsection{Results} \label{sec:expt-laypeople/results}

The retrieval models, namely BM25, SBERT and SAILER, perform poorly for laypeople queries. Their reliance on formal terms clashes with the informal language in queries, leading BM25 to miss matches and SAILER (trained on US/Canadian text) to fail in generalization. SBERT, though trained on general corpora, performs relatively better. The optimal $k$ values were 5, 3 and 6 for BM25, SBERT and SAILER respectively. Overall, retrieval methods remain weak baselines, compared to generative LLMs which surpass retrieval approaches even in ZSI. GPT-4.1 dominates across metrics, particularly recall (40.91\% $\mu$R in ZSI), despite no task-specific tuning. All models improve with TSI, but GPT-4.1 remains ahead in both precision and recall.  

Among open-source models, Llama-3 benefits most from supervised fine-tuning, reaching $\mu$F1 of 31.52\%, but performs poorly with RAG, indicating sensitivity to noisy retrieval. Gemma-3 is more balanced; TSI boosts its mF1, RAG lifts $\mu$R (31.53\%), and SFT yields its strongest overall trade-off ($\mu$F1 33.19\%). Thus, Gemma approaches GPT-4.1 on frequent statutes when fine-tuned, while Llama relies heavily on SFT to stay competitive.  

Interestingly, RAG underperforms TSI across all models, suggesting that a small number of in-context examples is more effective than retrieved statutes. This is largely due to limitations of the SBERT retriever used in RAG (for detailed analysis, see App .~\ref{app:sbert-analysis}). Comparing TSI and SFT, the former favors rare statutes (higher mF1), while the latter biases toward frequent ones (higher $\mu$F1). GPT-4.1 is unique in maintaining smoother performance across both regimes, underscoring its generalization ability even without fine-tuning.  

In summary, \textit{GPT-4.1 performs the best off-the-shelf, while Gemma-3 offers the most reliable open-source alternative when fine-tuned, and Llama-3 improves substantially only with SFT}.



\subsection{Analysis} \label{sec:expt-laypeople/analysis}
To make a more fine-grained analysis of the results in Table~\ref{tab:laypeople-results}, we analyze the effect of query category and statute frequency on the $\mu$F1 scores of the best performing models, i.e., GPT-4.1 (ZSI and TSI), and the SFT versions of Llama-3 and Gemma-3.

\begin{figure}[htbp]
  \centering
  \includegraphics[width=\linewidth]{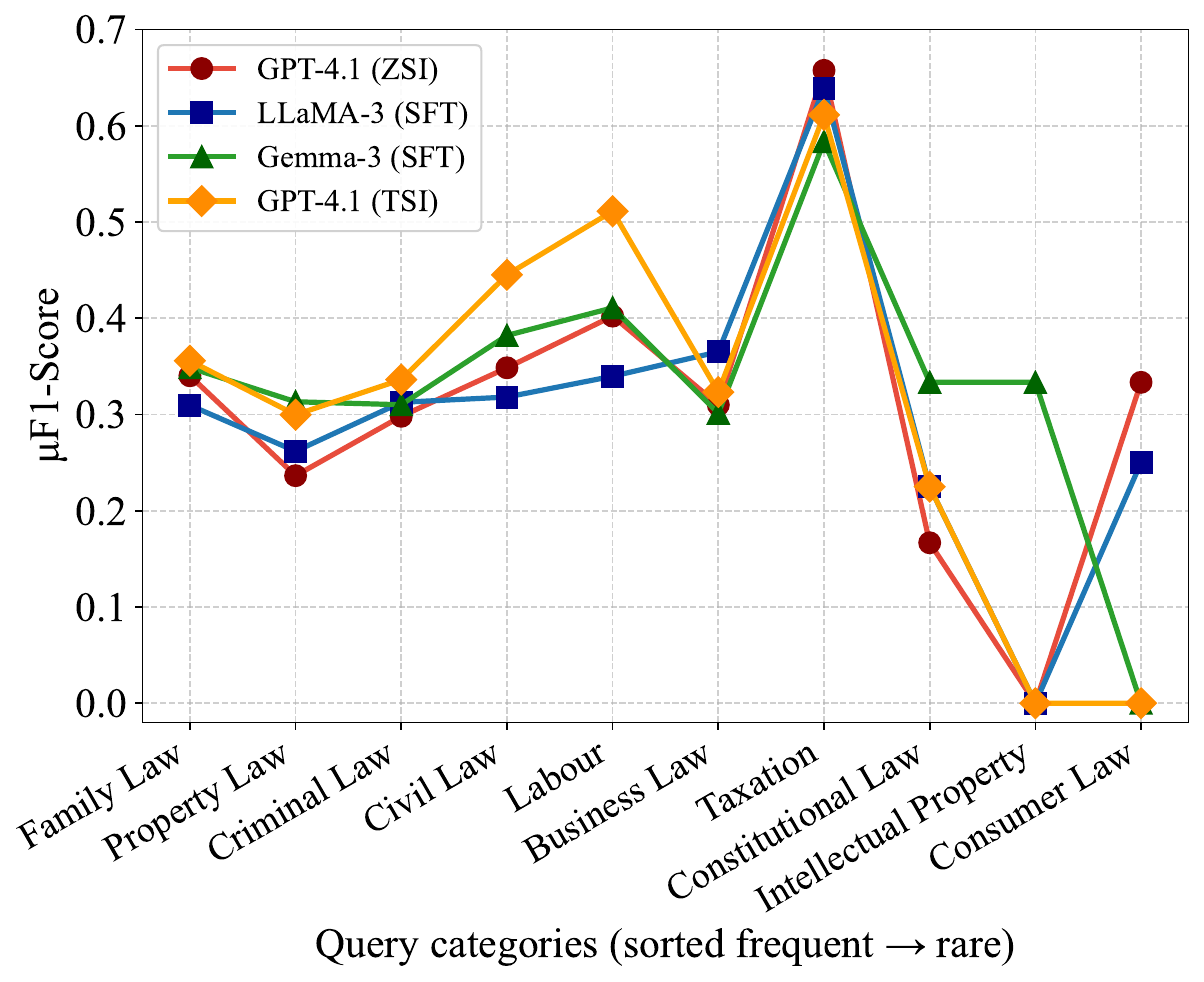}
  \caption{F1 Scores vs query categories sorted according to frequency over \tesuf{\ilsiclay} (left to right in descending order of category frequency)}
    \label{fig:laypeople-cat}
\end{figure}

\vspace{2mm}
\noindent \underline{\textbf{Effect of Query Category:}} We divide all the queries in \tesuf{\ilsiclay} into their respective categories, and calculate the $\mu$F1 score over each individual category. These results are plotted in Figure~\ref{fig:laypeople-cat}, where the X-axis represents each category, and the categories are sorted left to right in descending order of frequency over \tesuf{\ilsiclay}.

\noindent Scores generally decline for rarer categories such as Constitutional, I.P., or Consumer Law. 
An exception is \textit{Taxation}, where all models achieve unusually high $\mu$F1 ($\sim$0.6 vs. $\sim$0.4). 
This arises from data distribution: of six Taxation statutes in the test set, five also appear in training, leaving only one held-out statute, unlike adjacent categories (e.g., Business and Constitutional Law with 8 and 3 held-out statutes). 
Thus, the spike reflects overlap in statute coverage rather than true model capability.

\begin{figure}[!t]
  \centering
  \includegraphics[width=\linewidth]{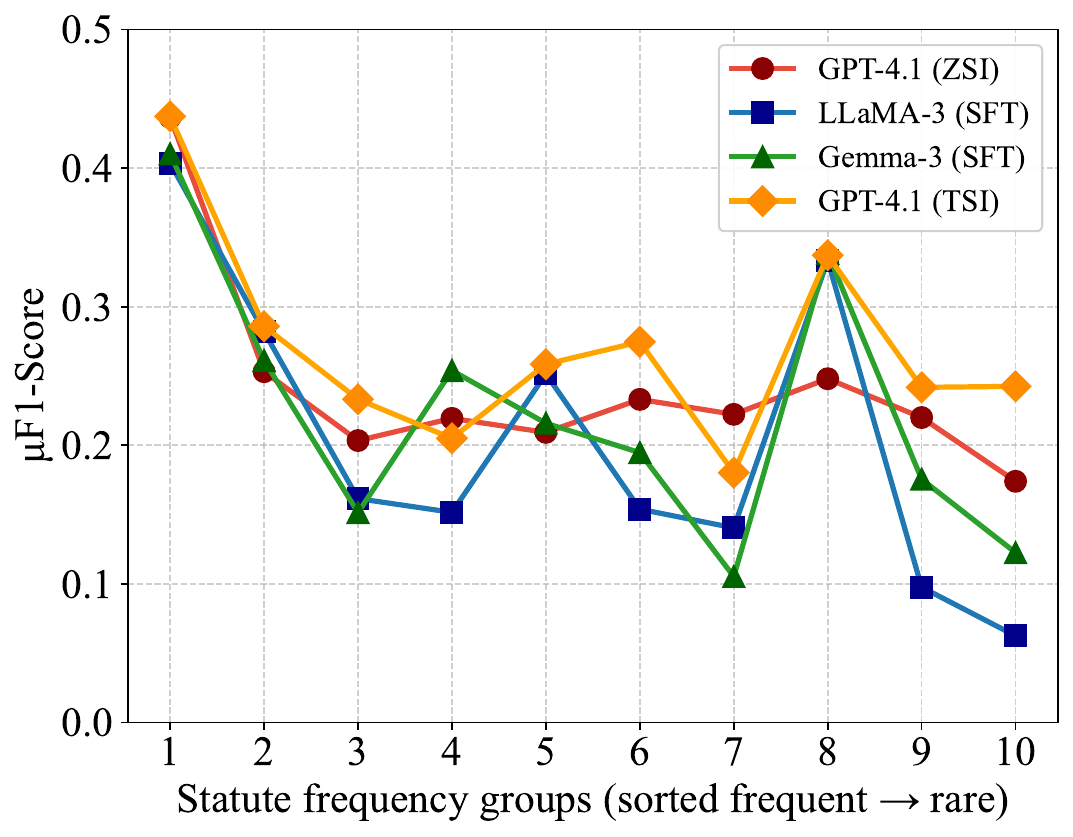}
  \caption{F1 Scores vs Statutes sorted according to frequency over \tesuf{\ilsiclay} and divided into 10 groups (left to right in descending order of statute frequency)}
    \label{fig:laypeople-freq}
\end{figure}

Across categories, Gemma generally outperforms Llama (notably in \textit{Constitutional} and \textit{I.P.} Law), except in \textit{Business}, \textit{Taxation}, and \textit{Consumer} Law, while GPT-4.1 tracks closely with either model depending on the category, showing particular gains in frequent categories under TSI.

\vspace{1mm}
\noindent \underline{\textbf{Effect of Statute Frequency:}} We sort all the statutes cited in the gold-standard of \tesuf{\ilsiclay}, arrange them into 10 groups and calculate the $\mu$F1 score over each individual group of statutes. These results are plotted in Figure~\ref{fig:laypeople-freq}, where the X-axis represents each group, and the statute groups are sorted left to right in descending order of frequency over \tesuf{\ilsiclay}.
Performance decreases as statutes become rarer, except for Group-8, which shows a sudden spike. These statutes mostly align with frequent categories (including Taxation), enabling models to exploit indirect category--frequency correlations. Fine-tuned models amplify this effect (steep Group-8 jump), while GPT-4.1 exhibits smoother performance trends, matching fine-tuned models on frequent Groups 1--5 but clearly surpassing them on rare Groups 6--10 (except Group-8), with TSI giving it a further edge on the rarest statutes.  

\vspace{1mm}
\noindent
\underline{\bf Performance on held-out statutes:}
As described in Section~\ref{sec:corpus/laypeople}, some statutes in the test set are held-out, i.e., only cited by test set queries, and not in train/dev sets. We test the generalization capabilities of the different models by calculating the metrics over the held-out statutes, and observe that GPT-4.1 still outperforms all other methods, while Gemma performs better than Llama in most settings (for more details, see App.~\ref{app:heldout}).

\vspace{1mm}
\noindent
\underline{\bf Ablation study on RAG pipeline:}
We conduct a set of ablation experiments on the RAG pipeline by varying the retrieval source, contextual information, and evidence count, and observe that RAG performance is sensitive to these design choices (details in App.~\ref{app:rag_ablation}).

\vspace{1mm}
\noindent
\underline{\bf Qualitative error analysis:}
We analyze common failure modes across ZSI, TSI, and SFT using representative examples (see App.~\ref{app:error_analysis}).

\begin{table}
    \small
    \centering
    \addtolength{\tabcolsep}{-0.28em}
    \begin{tabular}{l|cc|cc|cc}\toprule
         \multirow{2}{*}{\textbf{Model}} & 
         \multicolumn{2}{c|}{\textbf{ILSiC-GT}} & 
         \multicolumn{2}{c|}{\textbf{Annotated-GT}} & 
         \multicolumn{2}{c}{\textbf{ILSiC-A-GT}} \\
         & $\mathbf m$\textbf{F1} & $\mathbf\mu$\textbf{F1} & $\mathbf m$\textbf{F1} & $\mathbf\mu$\textbf{F1} & $\mathbf m$\textbf{F1} & $\mathbf\mu$\textbf{F1} \\\midrule
         \multicolumn{7}{c}{\textbf{Zero Shot Inference}} \\ \midrule
         Llama-3   & 1.74 & 19.33 & 1.56 & 11.49 & 0.79 & 9.58 \\
         Gemma-3  & 1.82 & 22.02 & 1.76 & 13.46 & 1.02 & 13.07 \\
         GPT-4.1 & \textbf{3.88} & 28.97 & \textbf{4.86} & \textbf{22.71} & \textbf{2.80} & \textbf{18.46} \\\midrule
         \multicolumn{7}{c}{\textbf{Supervised Fine-tuning}} \\ \midrule
         Llama-3 & 2.70 & 29.66 & 2.23 & 14.99 & 0.98 & 13.94 \\
         Gemma-3 & 2.47 & \textbf{32.88} & 2.17 & 14.58 & 1.13 & 17.00 \\\bottomrule
    \end{tabular}
    \caption{Performance (in \%) of base and fine-tuned models on human-annotated ground truth sets.}
    \label{tab:human-eval}
\end{table}

\subsection{Evaluation with Human Experts} 
To strengthen the reliability of the gold-standard annotations, we conducted a human evaluation with the help of an LL.M. graduate from a reputed Law school in India, hired as a subject matter expert for this project. The expert annotated relevant statutes for 50 randomly sampled queries. We compare three ground truth variants: \textbf{ILSiC-GT} (original dataset annotations), \textbf{Annotated-GT} (human expert annotations), and \textbf{ILSiC-A-GT }(their union). On average, ILSiC-GT contained 2.40 relevant statutes per query, Annotated-GT 5.46, and ILSiC-A-GT 6.53. This indicates a clear progression in coverage, where the original dataset captured only the most essential statutes, while human annotation included additional supporting ones. 


When evaluated against the human annotations, models showed slightly lower overall F1 scores due to a reduction in recall, as the expert annotated more statutes than the original dataset. However, \textit{the relative performance trends across the models were consistent} across all ground truth sets, suggesting that the dataset and the expert annotations are well in agreement. Even models fine-tuned on ILSIC continued to outperform base versions when tested on Annotated-GT and ILSiC-A-GT, underscoring that the improvements are not due to overfitting but reflect genuine generalization. Together, these findings demonstrate the robustness of our evaluation framework and dataset reliability.

\section{Experiments on \ilsicmulti} \label{sec:expt-multi/results}

The \ilsicmulti{} dataset enables a fair comparison of court vs.\ laypeople queries over the same set of statutes. We experiment with the following:

\vspace{1mm}
\noindent (i)~{\bf Zero-Shot Inference (ZSI):} Baseline inference.

\vspace{1mm}
\noindent (ii)~{\bf Laypeople Query Supervised Fine-tuning:} Models trained on \ltrsuf{\ilsicmulti} and validated on \lvdsuf{\ilsicmulti}, reflecting the test set.  

\vspace{1mm}
\noindent (iii)~{\bf Court Query Supervised Fine-tuning:} Models trained/validated on \ctrsuf{\ilsicmulti} and \cvdsuf{\ilsicmulti}, to test efficacy of court facts when evaluated on laypeople queries.  

\vspace{1mm}
\noindent (iv)~{\bf Synthetic Query Supervised Fine-tuning:} Court queries in \ctrsuf{\ilsicmulti} and \cvdsuf{\ilsicmulti} are re-written into laypeople style using GPT-4o-mini (Prompt in Table~\ref{tab:synthetic-prompt}), giving ${\ilsicmulti}_{train}^{syn}$ and ${\ilsicmulti}_{val}^{syn}$.  

\vspace{1mm}
\noindent (v)~{\bf Transfer Learning:} Sequential fine-tuning on \ctrsuf{\ilsicmulti}/\cvdsuf{\ilsicmulti} followed by \ltrsuf{\ilsicmulti}/\lvdsuf{\ilsicmulti}.  

\begin{table}
    \small
    \centering
    \addtolength{\tabcolsep}{-0.31em}
    \begin{tabular}{l|cc|cc|cc}\toprule
         \multirow{2}{*}{\textbf{Model}}&  \multicolumn{2}{c|}{\textbf{Precision}}&  \multicolumn{2}{c|}{\textbf{Recall}}&  \multicolumn{2}{c}{\textbf{F1}}\\
 & \textit{\bf m}& $\mu$& \textit{\bf m}& $\mu$& \textit{\bf m}&$\mu$\\\midrule
         \multicolumn{7}{c}{\textbf{Zero Shot Inference}}\\ \midrule
         GPT-4.1&  \textbf{16.71}&  28.53&  \textbf{21.93}&  \textbf{41.92}&  \textbf{17.01}& 33.95\\
         Llama-3&  7.92&  25.19&  6.04&  17.62&  5.64& 20.74\\
         Gemma-3&  10.45&  28.10&  7.58&  23.49&  7.35& 25.59\\ \midrule
         \multicolumn{7}{c}{\textbf{Supervised Fine-tuning on \ltrsuf{\ilsicmulti}}}\\ \midrule
         Llama-3&  12.67&  29.11&  14.74&  36.50&  12.21& 32.39\\
         Gemma-3& 12.74& \textbf{36.87}& 13.29& 32.88& 11.96&\textbf{34.76}\\ \midrule
         \multicolumn{7}{c}{\textbf{Supervised Fine-tuning on \ctrsuf{\ilsicmulti}}}\\ \midrule
         Llama-3& 7.27& 17.62& 10.52& 22.98& 6.52&19.95\\
         Gemma-3&  9.74&  10.42&  10.42&  23.38&  8.05& 23.10\\ \midrule
         \multicolumn{7}{c}{\textbf{Supervised Fine-tuning on }$\mathbf{{\ilsicmulti}_{train}^{syn}}$}\\ \midrule
         Llama-3& 8.05& 16.74& 10.55& 22.52& 7.11&19.20\\
         Gemma-3& 8.71& 19.78& 10.25& 23.21& 7.50&21.36\\ \midrule
         \multicolumn{7}{c}{\textbf{Transfer learn (\ctrsuf{\ilsicmulti} $\rightarrow$ \ltrsuf{\ilsicmulti})}}\\ \midrule
         Llama-3& 13.62& 29.87& 15.52& 36.56& 13.06&32.88\\
         Gemma-3& 12.03& 36.05& 13.00& 33.40& 10.99&34.67\\ \bottomrule
    \end{tabular}
    \caption{Performance of all methods on the \ilsicmulti{} dataset. All results are in percentage, evaluated on \tesuf{\ilsicmulti}.}
    \label{tab:multi-results}
\end{table}

\subsection{Results} \label{sec:expt-multi/results}
Table~\ref{tab:multi-results} shows results of different experiments on \ilsicmulti{}. The trends mirror the results over \ilsiclay{} ( Table~\ref{tab:laypeople-results}), with GPT-4.1 yielding the best mF1 under ZSI, but fine-tuned Llama-3 and Gemma-3 surpass it in $\mu$F1. 
Court-only training is ineffective, giving worse $\mu$F1 than ZSI and only marginal mF1 gains, confirming the linguistic mismatch between court and laypeople queries. 
Synthetic laypeople queries behave similarly, failing to capture laypeople-specific language. Transfer learning improves Llama-3 (best Llama result overall) but gives Gemma-3 only modest gains.

\subsection{Analysis} \label{sec:expt-multi/analysis}
We repeat the analyses of Section~\ref{sec:expt-laypeople/analysis} for query categories and statute frequency. Pure laypeople training consistently outperforms pure court training, while transfer learning results fluctuate (see App.~\ref{app:expt/multi-analysis}).



\section{Conclusion and Future Work} \label{sec:future}
In this resource paper, we create a new corpus \textbf{\texttt{ILSIC}} for identifying Indian legal statutes from laypeople queries. 
This is the first corpus that allows researchers to experiment with direct comparison between court facts vs. laypeople queries for LSI over the same set of statutes (\ilsicmulti{} dataset).
Our benchmarking over the \ilsiclay{} dataset shows that fine-tuned Llama and Gemma models are able to somewhat match the ZSI performance of GPT-4.1.
In the comparison experiments over \ilsicmulti{}, we observe that training purely over a large court dataset is completely ineffective for inference over laypeople queries, as compared to training on a smaller laypeople dataset. Transfer learning from court to laypeople data can help in certain cases, but there are still a lot of gaps to be addressed in terms of the relationship between court and laypeople data for LSI.
In future, we would like to try out other ways of combining the court and laypeople training data to further improve upon training purely on a small laypeople dataset.

\section*{Acknowledgments}

We thank the anonymous reviewers and the metareviewer for their insightful comments and suggestions. 
We also thank the law expert \textit{Shubham Tiwari}, Master of Laws (LLM) from the Rajiv Gandhi School Of Intellectual Property Law, India, who helped us conduct the annotation study. 

This research is
partially supported by the IIT Mandi iHub and HCi Foundation (iHub) (through the project titled ``Large Language Model for Legal Assistance'') and by the IIT Kharagpur Technology Innovation Hub on AI for Interdisciplinary Cyber-Physical Systems (AI4ICPS) (through the project titled ``NyayKosh: Multilingual Resources for AI- based Legal Analytics'').


\section*{Limitations}

In this paper, we created a new dataset for LSI over laypeople queries in the Indian legal scenario. 
We are the first, to the best of our knowledge, to create a dataset that allows direct comparison between using court facts vs. laypeople queries for LSI.
However, there are a few limitations of our study, which we describe briefly in this section.

Since most legal forums in India do not provide public access to question-answer data, we were able to collect laypeople queries from only a single website. 
Also, the dataset is limited to the English language, and does not cover the high linguistic diversity of India. 
It remains an important future work to develop a multilingual dataset of laypeople queries in various Indian languages.

For benchmarking on the \ilsiclay{} dataset, we have considered a limited number of models. This limited model coverage is partly due to hardware constraints. All experiments are conducted on a single 48GB Nvidia RTX A6000 GPU, which restricts evaluation to models of moderate size and prevents full fine-tuning of larger models.
In future, there is a need to experiment with more legal retrieval-focused models.


For comparing between court facts and laypeople queries for LSI, we have experimented with a limited number of settings. In terms of combining court and laypeople data for training, we have only utilized transfer learning, which produces marginal benefits over pure laypeople training only in limited cases. This aspect of the work demands further exploration by using other techniques to combine the two types of training data.

\section*{Ethical Considerations}

\noindent In this work, we develop a dataset that allows for the statute identification from laypeople queries. This task is crucial for the legal domain, and systems for this task are likely to help legal professionals as well as common persons. 
The AI methods / models are designed only to provide relevant recommendations to the legal professionals, or preliminary guidance to laypeople users, and are \textit{not} expected to be integrated directly into the decision-making process of the judicial system. 

The dataset is constructed from \textit{publicly available} webpages on \url{https://kaanoon.com}, an online Indian law forum. 
We ensured to collect only those queries which have been kept public by the people who asked the queries. 
We did not collect any query that was made private by the person who asked the query. 
Also, we manually reviewed 50 documents and found almost no personal identifiers such as person-names in the queries. 
Location names are present, and have been retained for legal relevance (since applicable statutes often depend on the particular Indian state/region), ensuring privacy while preserving essential context. 

The dataset is made available specifically for academic, non-commercial research, under the the Creative Commons Attribution-NonCommercial-ShareAlike (CC BY-NC-SA 4.0) license. 


\bibliography{references}



\clearpage
\newpage

\appendix

\section*{Appendix}


\titlecontents{section}[18pt]{\vspace{0.05em}}{\contentslabel{1.5em}}{}
{\titlerule*[0.5pc]{.}\contentspage} 

\titlecontents{table}[0pt]{\vspace{0.05em}}{\contentslabel{1em}}{}
{\titlerule*[0.5pc]{.}\contentspage} 

\startcontents[appendix] 
\section*{Table of Contents} 
\printcontents[appendix]{section}{0}{\setcounter{tocdepth}{4}} 

\startlist[appendix]{lot} 
\section*{List of Tables} 
\printlist[appendix]{lot}{}{\setcounter{tocdepth}{1}} 

\startlist[appendix]{lof} 
\section*{List of Figures} 
\printlist[appendix]{lof}{}{\setcounter{tocdepth}{1}} 

\newpage

\section{Details of the \texttt{ILSIC} Datasets} \label{app:dataset}

The \ilsiclay{} dataset comprises queries spanning a broad spectrum of legal categories, with separate splits for training, validation, and testing. It includes both commonly encountered topics such as Family Law and Property Law, as well as specialized areas like Intellectual Law and Taxation. The distribution of queries across categories reflects the diversity and prevalence of legal issues seen in real-world contexts, supporting the development and evaluation of robust legal language models. The detailed category-wise statistics are provided in \autoref{tab:ilsic-laypeople-category}

\begin{table}[ht]
\centering
\begin{tabular}{lccc}
\hline
\textbf{Category}         & \textbf{Train} & \textbf{Val} & \textbf{Test} \\
\hline
Family Law         & 2333  & 258 & 335  \\
Property Law       & 1924  & 251 & 223  \\
Criminal Law       & 1065  & 154 & 156  \\
Civil Law          & 645   & 80  & 52   \\
Labour             & 190   & 42  & 28   \\
Business Law       & 172   & 24  & 28   \\
Constitutional Law & 56    & 9   & 4    \\
Taxation           & 49    & 5   & 6    \\
Consumer Law       & 27    & 3   & 2    \\
Intellectual Law   & 4     & 0   & 2    \\
\hline
\textbf{Total}     & 6465  & 826 & 836  \\
\hline
\end{tabular}
\caption{Category-wise query distribution for the ILSIC laypeople dataset}
\label{tab:ilsic-laypeople-category}
\end{table}

\noindent
\textbf{Linguistic Differences between Laypeople and Court data:}
To understand the linguistic gap between the two domains, we conducted a detailed readability and lexical analysis of the laypeople and Court subsets in the ILSIC-Multi dataset. The results, summarized in \autoref{tab:readability_ilsic_multi}, reveal distinct stylistic and structural contrasts between the two types of texts.

The laypeople subset displays a higher Flesch Reading Ease score (64.53) and a lower Flesch–Kincaid Grade Level (9.29), suggesting that these queries are easier to comprehend and written in a simpler linguistic style. In contrast, the Court subset has a lower Reading Ease (46.81) and a higher Grade Level (12.38), reflecting the formal and complex nature of judicial text.

Lexical diversity: Type–Token Ratio (TTR) captures the proportion of unique words to total words, serving as a measure of vocabulary variety. However, since TTR is sensitive to text length, we also report Herdan’s C, a length-normalized measure of lexical richness. The laypeople subset shows a higher raw TTR (0.0203 vs. 0.0081) owing to its shorter, conversational phrasing. Yet, the Herdan’s C values are similar (0.7237 vs. 0.7027), suggesting that while Court texts are more repetitive and formulaic, they maintain comparable normalized lexical diversity.

Sentence length distribution: Average sentence length is slightly higher in Court documents (20.67 words) compared to laypeople queries (18.28 words), reflecting the longer and syntactically denser constructions typical of formal legal writing. The laypeople subset, by contrast, contains shorter and more direct sentences, consistent with natural, query-like expressions.

Overall, these measures demonstrate that the laypeople subset is more readable, lexically varied, and concise, whereas the Court subset is denser, more technical, and syntactically complex. This linguistic divergence underscores the challenge for models to generalize effectively between lay and formal legal domains.

\begin{table}[t]
\centering
\small
\setlength{\tabcolsep}{5pt}
\renewcommand{\arraystretch}{1.1}
\begin{tabular}{lcc}
\toprule
\textbf{Metric} & \textbf{Laypeople} & \textbf{Court} \\
\midrule
Average Flesch Reading Ease & 64.53 & 46.81 \\
Average Flesch--Kincaid Grade Level & 9.29 & 12.38 \\
Lexical Diversity -- TTR & 0.0203 & 0.0081 \\
Lexical Diversity -- HerdanC & 0.7237 & 0.7027 \\
Sentence Length Distribution & 18.278 & 20.674 \\
\bottomrule
\end{tabular}
\caption{Readability statistics for the Laypeople and Court in ILSIC-Multi.}
\label{tab:readability_ilsic_multi}
\end{table}

\begin{table*}[h!]
\centering
\begin{tabular}{|p{0.99\textwidth}|}
\hline
You are a highly knowledgeable legal expert. Your task is to identify and extract all legal references, such as statutes\_2993, acts, sections, and legal cases, explicitly mentioned in the input text. Follow these instructions carefully. \newline

\textbf{Instructions:} \newline
1) Extract \textbf{only} legal statutes\_2993, acts, sections, and cases that are \textbf{explicitly mentioned} in the input text. Do not infer or assume any references. \newline
2) Categorize the extracted information into the following structures: \newline
\ \ \textbf{Sections}: Each entry must include both the section number and its explicitly stated act. \newline
\ \ \textbf{Cases}: List all legal cases explicitly mentioned, including citations if provided. \newline
3) Replace \texttt{"Indian Penal Code"} with \texttt{IPC} and \texttt{"Criminal Penal Code"} with \texttt{CrPC} wherever applicable. \newline
4) Ensure high accuracy and completeness. Do not omit or add any information beyond what is explicitly stated. \newline
5) Do not output acts separately. Each section must explicitly include its associated act. \newline
6) If multiple sections belong to the same act, list them as separate objects. \newline
7) If no references are found, return empty lists (\texttt{[]}). \newline
8) Handle partial or ambiguous references by including them as-is without interpretation. \newline

\textbf{Output Format (JSON only):} \newline
\texttt{\{} \newline
\ \ \texttt{"sections": [} \newline
\ \ \ \ \texttt{\{} \newline
\ \ \ \ \ \texttt{"act": "IPC or CrPC or other explicitly stated act",} \newline
\ \ \ \ \ \texttt{"section": "section number as mentioned"} \newline
\ \ \ \ \texttt{\}} \newline
\ \ \texttt{],} \newline
\ \ \texttt{"cases": [} \newline
\ \ \ \texttt{"list of explicitly mentioned cases"} \newline
\ \ \texttt{]} \newline
\texttt{\}} \newline

\textbf{Rules:} \newline
- Do not infer missing acts or sections. \newline
- Maintain official legal naming standards. \newline
- Output must strictly follow the specified JSON structure. \newline
- Do not include commentary, explanations, or extra text. \newline

\textbf{User Input:} \newline
\texttt{QUERY:} \newline
\{Target\_query\_text\} \newline

\textbf{Return only the JSON object.} \\
\hline
\end{tabular}
\caption{Prompt used for Legal Reference Extraction Task}
\label{tab:legal-extraction-prompt}
\end{table*}

\section{Hyper-parameters for Fine-tuning Experiments} \label{app:hyperparams}

For all fine-tuning experiments, we tried out for maximum of 5 epochs with early stopping. For SAILER, we used a batch size of 16, and 1 positive sample and 47 negative samples per query. SAILER was fine-tuned using Contrastive Loss with in-batch negative sampling. We used a l.r. 5e-6.
For LLMs, we used QLoRA for fine-tuning, with the LoRA parameters being $\alpha=32$, $r=8$, dropout at 0.05, l.r. 1e-5 and and 4-bit quantization. We used a batch size of 4, and trained using the standard Cross Entropy Loss. The maximum sequence length was kept at 32k. The LLMs were trained over \textit{completion only}.

\begin{table}[!t]
    \small
    \centering
    \addtolength{\tabcolsep}{-0.22em}
    \begin{tabular}{l|cc|cc|cc}\toprule
        \multirow{2}{*}{Model} & \multicolumn{2}{c|}{Precision} & \multicolumn{2}{c|}{Recall} & \multicolumn{2}{c}{F1 Score}\\
        & m & $\mu$ & m & $\mu$ & m & $\mu$\\\midrule
        \multicolumn{7}{c}{Zero Shot Inference (ZSI)}\\\midrule
        Llama-3 & 3.25 & 77.78 & 18.75 & 17.08 & 22.88 & 28.00\\
        Gemma-3 & 5.00 & 100.00 & 18.75 & 17.08 & 26.05 & 29.17\\
        GPT-4.1 & \textbf{6.25} & \textbf{100.00} & \textbf{38.55} & \textbf{36.59} & \textbf{44.05} & \textbf{53.58}\\\midrule
        \multicolumn{7}{c}{Two Shot Inference (TSI)}\\\midrule
        Llama-3 & 3.25 & 80.00 & 20.83 & 20.42 & 30.42 & 31.38\\
        Gemma-3 & 50.00 & 100.00 & 23.75 & 21.96 & 31.75 & 34.60\\
        GPT-4.1 & \textbf{6.25} & 100.00 & 30.21 & 27.78 & 40.38 & 46.29\\\midrule
        \multicolumn{7}{c}{Retrieval-Augmented Generation (RAG)}\\\midrule
        Llama-3 & 25.00 & 8.75 & 6.25 & 4.17 & 7.14 & 4.77\\
        Gemma-3 & 25.0 & 100.0 & 5.63 & 4.84 & 9.17 &9.41\\\midrule
        \multicolumn{7}{c}{Supervised Fine-Tuning (SFT)}\\\midrule
        Llama-3 & 12.5 & 100.0 & 7.29 & 7.22 & 11.31 & 9.17\\
        Gemma-3 & 12.5 & 100.0 & 5.00 & 4.88 & 7.15 &9.31\\\bottomrule
    \end{tabular}
    \vspace{-1mm}
    \caption{Evaluation across held-out statutes under different experimental setups.}
    \vspace{-6mm}
    \label{tab:heldout-results}
\end{table}

\section{Court and Laypeople Query Examples on the Same Statute}
\label{app:laypeople-court-examples}

To illustrate the differences in language and style between laypeople and court queries relating to the same statute, we present a representative pair of queries below. Each pair includes a query from a court document and a corresponding laypeople query, both referring to the same legal provision(s).

\vspace{0.7em}
\noindent
\textit{Example: Section 12 of The Consumer Protection Act, 1986 and Section 427 of The Indian Penal Code, 1860}

\begin{table*}[htbp]
\centering
\label{tab:court-query-consumer-ipc}
\begin{tabular}{|p{0.96\textwidth}|}
\hline
The present appeals arise out of the impugned judgment and order dated 26.04.2013 in Revision Petition No. 2032 of 2012 and order dated 23.07.2013 in Review Petition No. 253 of 2013 passed by the National Consumer Disputes Redressal Commission, New Delhi (hereinafter referred to as the National Commission), whereby the petitions challenging the order dated 29.02.2012 passed by the Haryana State Consumer Disputes Redressal Commission were dismissed. The brief facts of the case which are required to appreciate the rival legal contentions advanced by the learned counsel appearing on behalf of the parties are stated in brief as hereunder:- The appellant was the owner of a Tata Motors goods carrying vehicle bearing registration No.HR-67-7492. The vehicle was insured with the respondent- Company vide policy No. 15019923334104992 with effect from 31.07.2009, valid upto 30.07.2010. The risk covered in this policy was to the tune of Rs.2,21,153/-. The said vehicle met with an accident on 11.02.2010 on account of rash and negligent driving of the offending vehicle bearing registration no. UP-75-J 9860. In this regard, an FIR No.66 of 2010 dated 11.02.2010 was registered with the jurisdictional Police Station, Sadar, Fatehabad, for the offence punishable under [SECTION], [SECTION], [SECTION] and [SECTION] of the [ACT] (hereinafter referred to as the [ACT]). The appellant incurred expenses amounting to Rs.1,64,033/- for the repair of his vehicle and also informed the respondent- Company about the accident and damage caused to his vehicle. In this connection, the respondent- Company appointed one Mr. [ENTITY], as the Surveyor to assess the damage caused to the said vehicle. After inspecting the vehicle, the Surveyor assessed the damage caused to the vehicle at Rs.90,000/-, whereas the appellant had preferred a claim for a sum of Rs.1,64,033/- with supporting bills. In addition to above, the respondent-Company appointed M/s Innovation Auto Risk Claim Manager for the purpose of investigation. According to the report of the investigator, five passengers were travelling in the goods-carrying vehicle, though the seating capacity of the vehicle as per the registration certificate was only 1+1. On the basis of findings of the said report, the respondent-Company vide letter dated 26.07.2010 rejected the claim of the appellant for the reason that the loss did not fall within the scope and purview of the insurance policy. Aggrieved of the letter of rejection of the claim of the appellant by the respondent-Company, he filed Complaint No.517 of 2010 against the respondent-Company dated 17.09.2010 before the District Consumer Disputes Redressal Forum, Sonepat (hereinafter referred to as the District Forum) under [SECTION] of the [ACT] for the claim of Rs.1,64,033/- towards the repair of his vehicle on the ground that the rejection of the claim amounts to deficiency in service on the part of the respondent-Company. The respondent-Company filed a detailed written statement before the District Forum disputing the claim of the appellant. It took the plea that the complainant had violated the terms and conditions of the policy, as five passengers were travelling in the goods-carrying vehicle at the time of accident, whereas the permitted seating capacity of the motor vehicle of the appellant was only 1+1. The District Forum on the basis of the pleadings of the parties and the materials on record considered the judgment of the National Commission in the case of [PRECEDENT]], wherein it was held that if the number of persons travelling in the vehicle at the time of the accident did not have a bearing on the cause of accident, then the mere factum of the presence of more persons in the vehicle would not disentitle the insured claimant from claiming compensation under the policy towards the repair charges of the vehicle paid by the appellant. The District Forum accordingly directed the respondent-Company to settle the claim of the appellant on non-standard basis upto 75\% of the amount spent for effecting repairs to the damaged vehicle after taking into consideration the claim amount of Rs.1,64,033/-. The District Forum further directed the respondent-Company to settle the amount to be paid to the appellant along with interest at the rate of 9\% per annum from the date of lodging of the claim by the appellant with the respondent-Company. The respondent-Company was further directed to pay Rs.2,000/- for rendering deficient service, causing mental agony and harassment and towards litigation expenses incurred by the appellant. Aggrieved of the order of the District Forum, the respondent Company preferred an appeal before the State Commission urging various grounds.
\\\hline
\end{tabular}
\caption{Court Query on Consumer Protection and IPC Section 427}
\label{tab:court-query}
\end{table*}

\begin{table*}[ht]
\centering
\label{tab:laypeople-query-consumer-ipc}
\begin{tabular}{|p{0.96\textwidth}|}
\hline
I am a retired person and I wanted to build a roof top green house garden. As I don't know anyone who is in the field I searched online and found some companies in Indiamart website. I got several names and chose one (but no basis for my selection as the guy called me within 30min). Initially he was very polite and so thought he is nice person ... so I entered the contract with him. Initialy I paid him Rs 50000 (cheque) and he brought the structural materials next day and instructed his guys as how to the work. The work went well for a few days and subsequently all went wrong (meaning no work) as he kept giving excuses. While I gave a threatening call he sent another guy (he said his brother) and took up the work to complete the structural work sincerely. Subsequently, the same brother told me that we need to get the roof material and need Rs 50000. I said the guy who made contrct with me should tell and then I give, within a few mins the guy called and told me that you give 50000 cheque. As soon as they took money from bank both of their cell is off all time. Some times it rings but no response. I can afford the money though I am retired. But as a true citizen of India I want them to get punished. I would appreciate if you could suggest me to take forward to curb such criminal. Besides, your suggestion of a honest lawyer will help me to make these guys not to cheat anyone in their life. I live in whitefield area in Bangalore$>$ Thanks and I hope that I will get fruitful suggestion.
\\ \hline
\end{tabular}
\caption{Laypeople Query on Consumer Protection and IPC Section 427}
\label{tab:lay-query}
\end{table*}

\noindent
\textit{Description:} \\
The court query \autoref{tab:court-query} is highly structured, referencing procedural history, evidence, and legal claims in formal and technical language. In contrast, the laypeople query \autoref{tab:lay-query} tells a personal story in simple language, explains events chronologically, and focuses on seeking practical advice and justice, without legal terminology.

\begin{figure*}[h]
  \centering
  \begin{subfigure}[b]{0.48\textwidth}
    \centering
    \includegraphics[width=\textwidth]{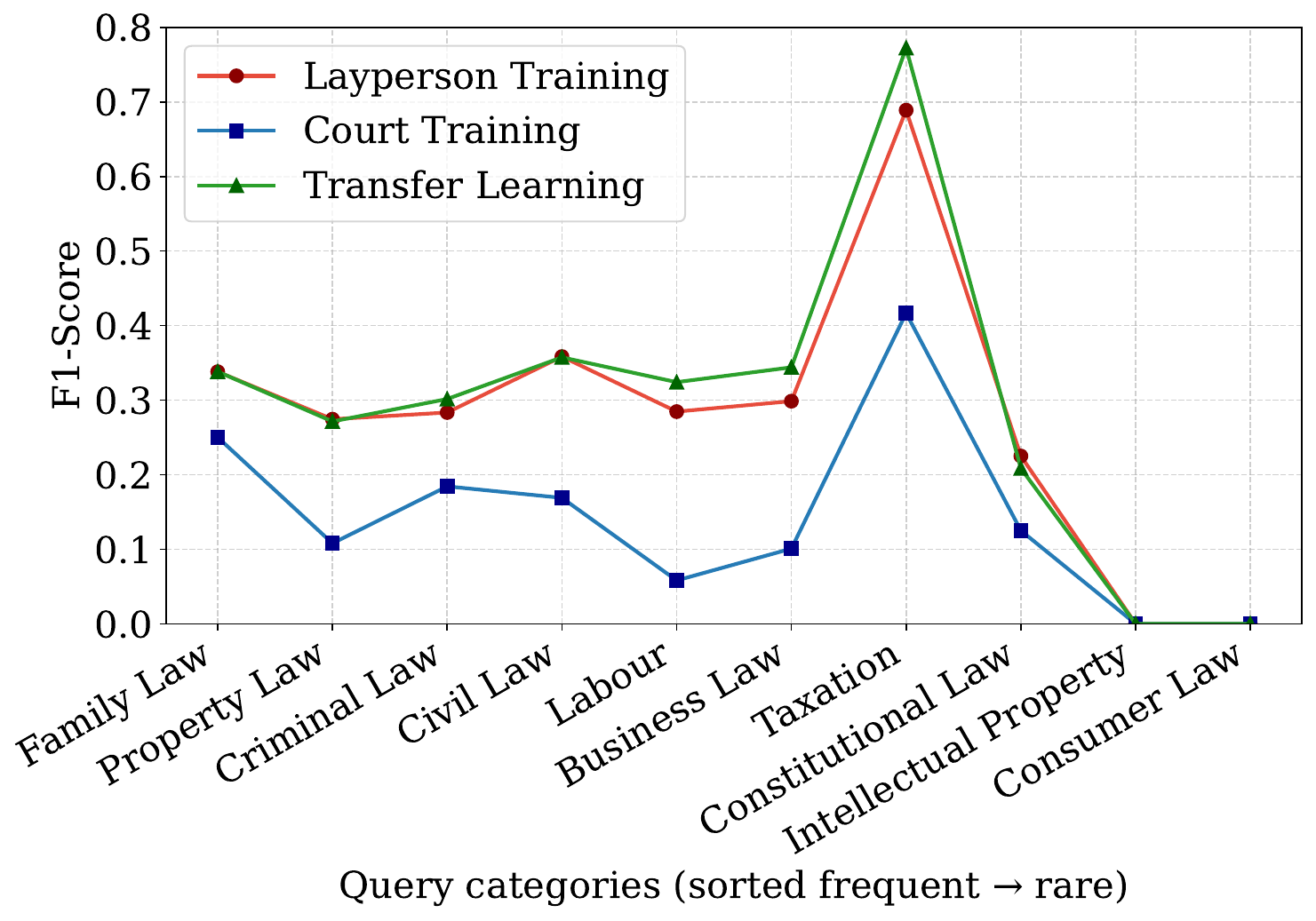}
    \caption{Llama-3}
    \label{fig:multi-cat-llama}
  \end{subfigure}
  \begin{subfigure}[b]{0.48\textwidth}
    \centering
    \includegraphics[width=\textwidth]{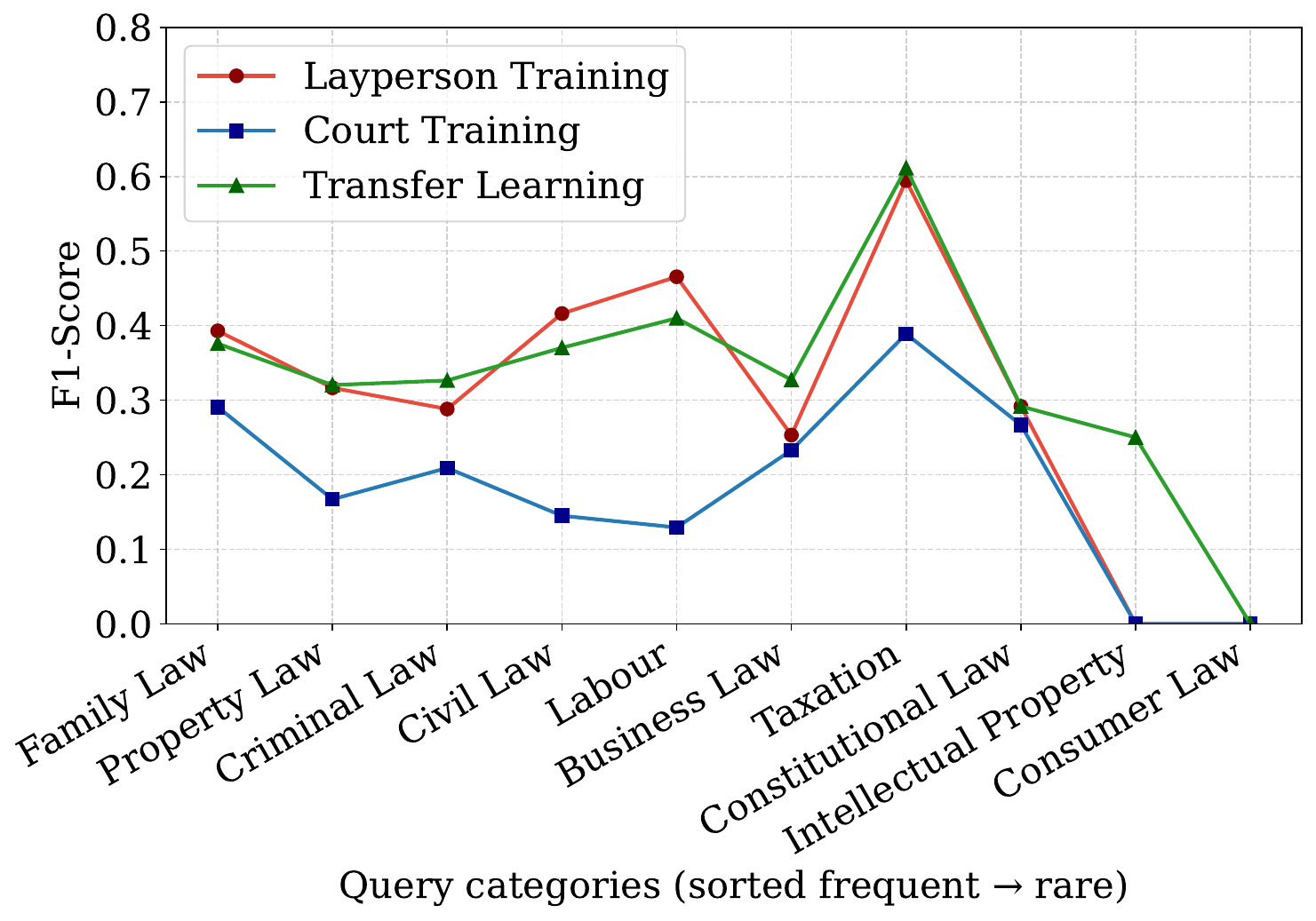}
    \caption{Gemma-3}
    \label{fig:multi-cat-gemma}
  \end{subfigure}
  \caption{Comparison of fine-tuning models across different settings and categories of queries}
  \vspace{-3mm}
  \label{fig:multi-cat}
\end{figure*}

\section{Verbalization Algorithm for LLM inference} \label{app:verb}

To ensure consistent and interpretable statute references during generation, we employ a structured verbalization pipeline that converts model outputs into standardized \textit{Section--Act} pairs. The process begins by isolating potential statute mentions from the generated text using regular expressions tuned to legal numbering patterns (e.g., ``Section~420 of the Indian Penal Code''). Each extracted mention is then normalized through lowercasing, removal of connective fillers (e.g., \textit{of}, \textit{under}, \textit{by}), and punctuation stripping. 

For robust alignment with canonical references, we perform hybrid matching that combines exact section matching with fuzzy act similarity computed via a weighted ensemble of RapidFuzz metrics, ratio, token-sort, and token-set, supplemented by a Jaccard word-overlap score and selective boosts for key legal terms. A match is accepted only when the section identifier matches exactly and the act-name similarity exceeds a confidence threshold ($\tau \approx 70$--$85$). 

This verbalization step harmonizes minor surface variations (e.g., ``IPC''~$\leftrightarrow$~``Indian Penal Code'', ``Motor Vehicles Act''~$\leftrightarrow$~``MV Act''), yielding consistent statute identifiers across LLM outputs. On a manually validated sample of 50 generation instances, the algorithm achieved approximately 95\% correct statute normalization, with most residual discrepancies arising from irregular act abbreviations or nested references within multi-clause sections.

\begin{figure*}[htbp]
  \centering
  \begin{subfigure}[b]{0.48\textwidth}
    \centering
    \includegraphics[width=\textwidth]{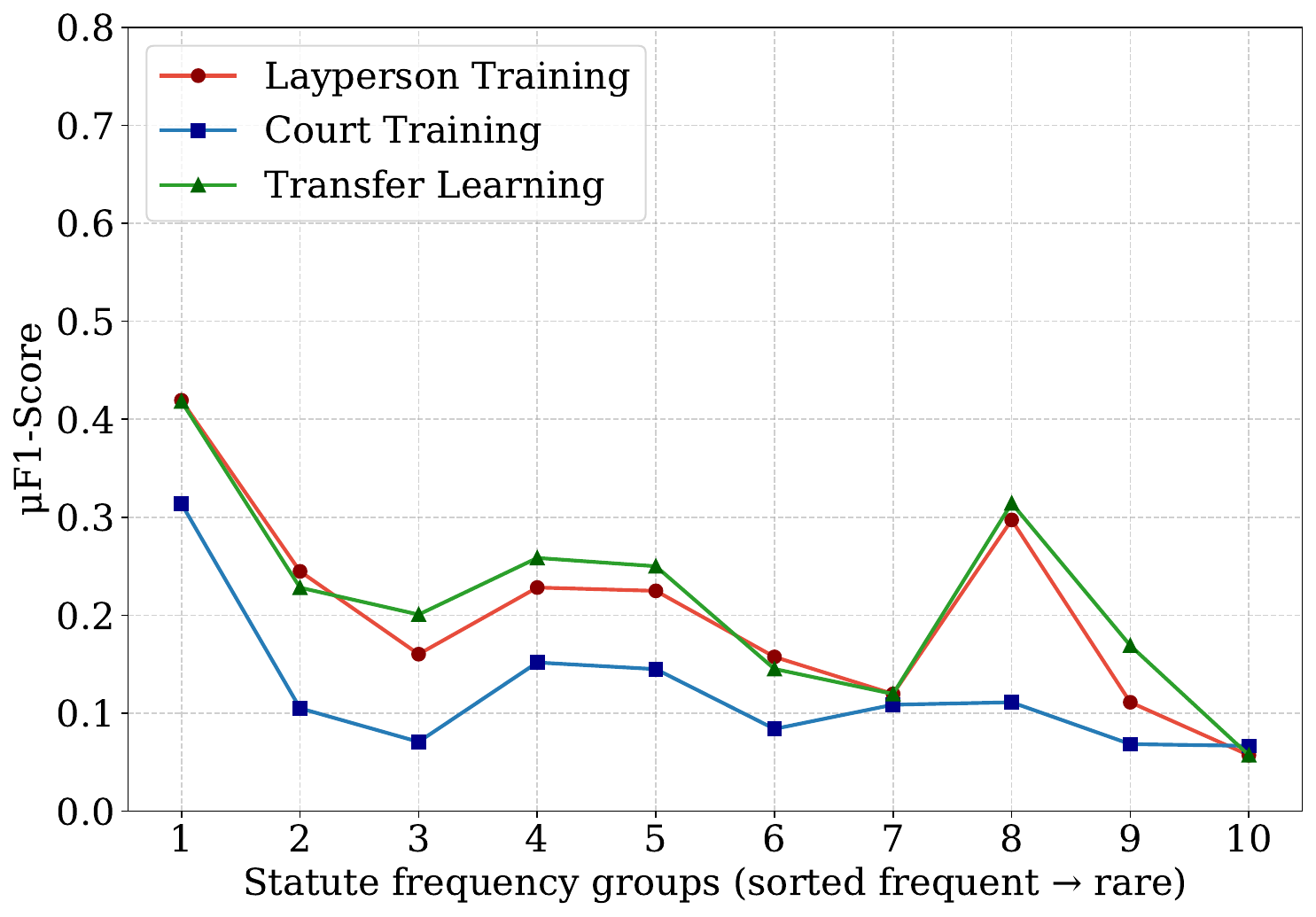}
    \caption{Llama-3}
    \label{fig:multi-freq-llama}
  \end{subfigure}
  \begin{subfigure}[b]{0.48\textwidth}
    \centering
    \includegraphics[width=\textwidth]{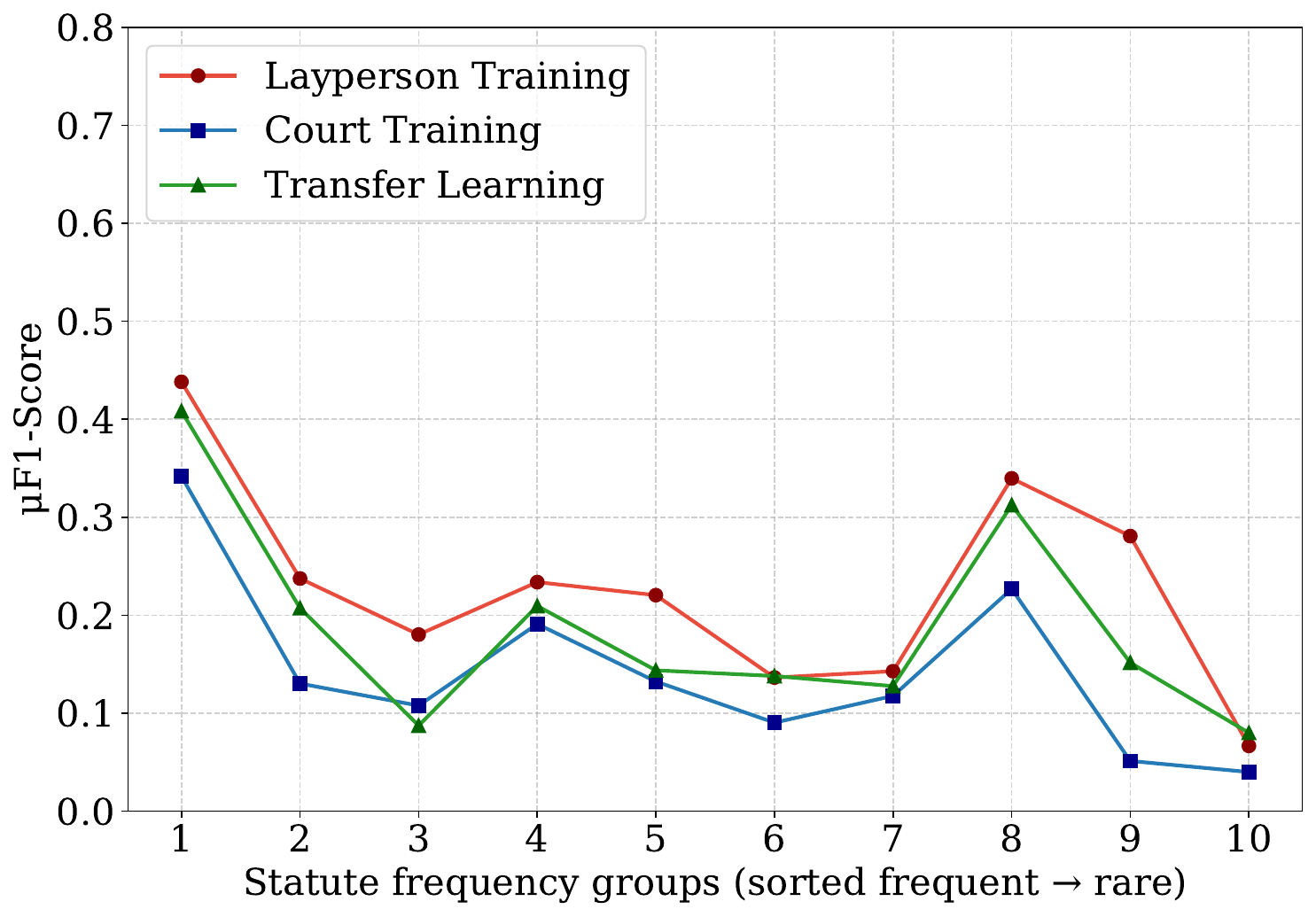}
    \caption{Gemma-3}
    \label{fig:multi-freq-gemma}
  \end{subfigure}
  \caption{Comparison of fine-tuning models across different settings and frequency of statutes}
  \label{fig:multi-freq}  
  \vspace{-3mm}
\end{figure*}

\vspace{1em}
\section{Prompts used for LLM Experiments} \label{app:prompts}
The prompt shown in \autoref{tab:rag-sbert-prompt} serves as a foundational component in our approach to statutory retrieval and model fine-tuning. By simulating the role of a legal expert, the prompt provides clear and restrictive instructions for the extraction of statutory references, requiring the identification of precise section and act pairs directly relevant to a given legal scenario. This structure not only ensures the completeness and accuracy of extracted information but also standardizes output for consistent model training and evaluation. As a result, the prompt enables effective fine-tuning of language models on legal tasks, supports robust retrieval of statutes from complex legal texts, and lays the groundwork for building practical AI tools for legal research, compliance checking, and legal decision support in the Indian legal context.
\begin{table*}
    \centering
    \begin{tabular}{|p{0.99\textwidth}|}
    \hline
    You are a legal expert on Indian law. Given the text below, identify all applicable legal sections and their full act names. Follow these rules exactly: \newline
        1. Identify every relevant provision (section number + full act name) you are 100\% certain applies under Indian statutes. \newline
        2. Output only one list in square brackets, formatted like: ["Section X of Act Name"; "Section Y of Act Name"; …]\newline
           - Each entry must be a quoted string with the section first, then the full act name.\newline
            - Each entry needs both section number AND full act name\newline
            - Separate entries with a semicolon.\newline
            - Do not include anything outside this single list.\newline
        3. Exclude any entry missing either section or act (no incomplete pairs).\newline
        4. Do not repeat identical section-act pairs (no duplicates).\newline
        5. Do not add explanations, labels, or extra-only the list itself.\newline
        6. Always spell out the full name of each act (no abbreviations).\newline
        7. Include only provisions that are clearly and directly relevant (no speculative or uncertain entries).\\ \hline
    \end{tabular}
    \caption{Prompt used for Zero-Shot Inference and Supervised Fine-Tuning Task}
    \label{tab:llm-prompt}
\end{table*}

The prompt detailed in \autoref{tab:synthetic-prompt} is designed to guide the model in generating clear and accessible summaries of Indian court cases for non-expert audiences. By instructing the model to avoid legal jargon, technical references, or placeholder terms, and to present the facts in a conversational and easily understandable manner, the prompt ensures the creation of realistic laypeople-style case summaries. This approach is essential for constructing synthetic datasets that reflect how actual litigants or the general public might describe their legal issues online. Such datasets are crucial for training and evaluating models intended for laypeople-facing legal applications.

\begin{table*}[htbp]
    \centering
    \begin{tabular}{|p{0.99\textwidth}|}
    \hline
    You are a legal expert specializing in Indian court cases. I will provide you with the facts of a court case. Your task is to generate a laypeople's summary of the case, written in a single, continuous paragraph, without using bullet points, numbered lists, or section headers. The summary should be clear, concise, and easy to understand for someone without legal expertise, but it must include all relevant facts and details. Write in a conversational style, as if you are personally describing your situation and seeking advice, similar to how a layperson might post their question on an online forum. \newline
    \newline
    Important: Do not label or mention legal entities, sections, acts, or precedents with placeholders such as [ENTITY], [SECTION], [ACT], [PRECEDENT], etc. Do not add, invent, or fill in any legal names, sections, or references that are not present in the original case facts I provide. If the court facts I provide do not mention a statute (such as a section or act), you must not include any such statute in your output, under any circumstance. Only summarize the information actually present in the input.\newline
    \newline
    Only output the summary in the required style.\\ \hline
    \end{tabular}
    \caption{Prompt used for Synthetic laypeople dataset generation}
    \label{tab:synthetic-prompt}
\end{table*}

The prompt shown in \autoref{tab:two-shot-template} corresponds to the \textit{two-shot inference setup}, where the model is guided using two semantically similar example pairs before generating the output for the target query. Each example provides a \texttt{query-text} and its corresponding list of statute predictions, enabling the model to observe the expected input–output structure. This few-shot formulation helps the model generalize the pattern of statute identification by leveraging contextual cues from analogous cases.
.
\begin{table*}[h]
    \centering
    \begin{tabular}{|p{0.99\textwidth}|}
    \hline
    You are a legal expert on Indian law. Given the text below, identify all applicable legal sections and their full act names. Follow these rules exactly: \newline
        1. Identify every relevant provision (section number + full act name) you are 100\% certain applies under Indian statutes. \newline
        2. Output only one list in square brackets, formatted like: ["Section X of Act Name"; "Section Y of Act Name"; …]\newline
           - Each entry must be a quoted string with the section first, then the full act name.\newline
           - Each entry needs both section number AND full act name.\newline
           - Separate entries with a semicolon.\newline
           - Do not include anything outside this single list.\newline
        3. Exclude any entry missing either section or act (no incomplete pairs).\newline
        4. Do not repeat identical section–act pairs (no duplicates).\newline
        5. Do not add explanations, labels, or extra text-only the list itself.\newline
        6. Below are two EXAMPLES of the task (input~1/2 with output1/2). Then produce the answer ONLY for input~3 by writing output3. Stop immediately after you close the single list with a `]`. \newline
        \newline
        \textbf{input 1:} \{Example\_1\_query\_text\}\newline
        \textbf{query-religion:} \{Religion\_1\} \newline
        \textbf{output1:} ["Section ... of Act ..."; "Section ... of Act ..."] \newline
        \newline
        \textbf{input 2:} \{Example\_2\_query\_text\}\newline
        \textbf{query-religion:} \{Religion\_2\} \newline
        \textbf{output2:} ["Section ... of Act ..."; "Section ... of Act ..."] \newline
        \newline
        \textbf{input 3:} \{Target\_query\_text\}\newline
        \textbf{query-religion:} \{Religion\_3\} \newline
        \textbf{output3:} [Model to generate this list]\newline
        \\ \hline
    \end{tabular}
    \caption{Prompt used for Two-Shot Inference Task}
    \label{tab:two-shot-template}
\end{table*}

The prompt shown in \autoref{tab:rag-sbert-prompt} is designed for the \textit{retrieval-augmented generation (RAG)} stage. Here, the model receives a user query along with a set of candidate statutes retrieved via SBERT similarity and their formal definitions. The model is explicitly instructed to select only the most relevant statutes from this constrained set, adhering strictly to the given candidates without rephrasing or inventing new entries. This structured prompt ensures interpretability with the retrieved legal context, thereby improving reliability in downstream re-ranking and selection tasks.

\begin{table*}[h]
    \centering
    \begin{tabular}{|p{0.99\textwidth}|}
    \hline
    You are a legal expert on Indian law. Select statutes that best apply to the user's situation, but \textbf{ONLY} from the provided \texttt{CANDIDATE\_STATUTES}. \newline
    \textbf{Rules:} \newline
    1) Output exactly one list in square brackets using double quotes, e.g., \texttt{["Section X of Act"; "Section Y of Act"]}. \newline
    2) Each entry must be a quoted string in the \emph{exact same form} as provided in \texttt{CANDIDATE\_STATUTES} (no rephrasing). \newline
    3) Do not add anything outside that one list. No commentary. \newline
    4) Do not invent or add statutes not in \texttt{CANDIDATE\_STATUTES}. \newline
    5) Prefer fewer, more precise provisions over many generic ones. \newline
    6) If none clearly apply, return \texttt{[]}. \newline
    \newline
    \textbf{User message (per query).} \newline
    \texttt{QUERY:} \newline
    \{Target\_query\_text\} \newline
    \newline
    \texttt{CANDIDATE\_STATUTES:} \newline
    \ \ \texttt{- "Section \dots\ of Act \dots"} \newline
    \ \ \texttt{- "Section \dots\ of Act \dots"} \newline
    \ \ \texttt{- \dots} \newline
    \newline
    \texttt{STATUTE\_DEFINITIONS:} \newline
    \ \ \texttt{- "Section \dots\ of Act \dots": \{definition text\}} \newline
    \ \ \texttt{- "Section \dots\ of Act \dots": \{definition text\}} \newline
    \ \ \texttt{- \dots} \newline
    \newline
    \texttt{Return only the bracketed list.} \\ \hline
    \end{tabular}
    \caption{Prompt used for RAG Task}
    \label{tab:rag-sbert-prompt}
\end{table*}

\section{Details of Experiments on \ilsicmulti{}} \label{app:expt/multi-analysis}

Since our objective in this Section is to compare the effects of different fine-tuning data, we plot only the results for pure laypeople training, pure court training and transfer learning for both Llama and Gemma.

\noindent \underline{\textbf{Effect of Query Category:}} The performance distribution with categories for both models are shown in Figure~\ref{fig:multi-cat}. 
We observe the same spike in performance for \textit{Taxation} for both models as seen in Figure~\ref{fig:laypeople-cat}. For Llama-3, court training significantly underperforms for all categories, while transfer learning performs marginally better as compared to laypeople training (see Figure~\ref{fig:multi-cat-llama}). Similar trends are observed for Gemma-3 as well (see Figure~\ref{fig:multi-cat-gemma}), however, transfer learning clearly underperforms compared to pure laypeople training for two categories, \textit{Civil} and \textit{Labour} Law.

\noindent \underline{\textbf{Effect of Statute Frequency:}} The performance distribution across frequency of statutes are shown in Figure~\ref{fig:multi-freq}. 
Again, similar to Figure~\ref{fig:laypeople-freq}, we observe a spike in performance for both models on Group-8.
For Llama-3, transfer learning performance is similar (but slightly less) to pure laypeople training across all groups, while court training underperforms.
For Gemma-3, however, transfer learning performance reduces to the level of pure court training for many of the frequent groups (Groups 3-5).

\section{Performance on Held-Out Statutes} \label{app:heldout}
To assess the models’ generalization beyond the training distribution, we evaluated their performance on a subset of $8$ held-out statutes and the corresponding $41$ queries that referenced them. This analysis captures the models’ ability to reason over unseen statutory provisions rather than relying on memorized associations. As shown in \autoref{tab:heldout-results}, the zero-shot inference (ZSI) and two-shot inference (TSI) settings demonstrated the strongest results among the evaluated setups. In particular, GPT-4.1 achieved the highest overall scores, with macro- and micro-F1 values of 0.4475 and 0.5358 in the zero-shot setting, reflecting consistent generalization to unfamiliar statutes. Gemma and Llama 3 models followed with moderate gains, especially under the two-shot configuration, indicating that limited in-context exemplars improve adaptation to unseen sections.

In contrast, the retrieval-augmented generation (RAG) and supervised fine-tuning (SFT) approaches exhibited lower precision and recall, suggesting that exposure to retrieved or previously seen statutes did not transfer effectively to unseen legal provisions. These results emphasize that the generative reasoning capacity of large language models, particularly GPT-4.1, plays a more decisive role in extrapolating to new statutes than retrieval or task-specific fine-tuning. Overall, the held-out evaluation highlights both the challenge of statute generalization and the potential of few-shot prompting to mitigate it.

\section{Analysis of SBERT Retrieval Performance} \label{app:sbert-analysis}
We evaluate the effectiveness of SBERT as the retriever in our RAG pipeline by measuring its ability to retrieve the ground-truth statute within the top-15 candidates. As shown in Table~\ref{tab:sbert_rag_analysis}, SBERT achieves a macro precision of only 31.99\%, indicating limited retrieval quality.

This retrieval bottleneck directly affects the downstream RAG performance, as the language model is conditioned on a noisy and incomplete candidate set. Consequently, RAG does not yield improvements over Zero-Shot Inference (ZSI) and, in some cases, performs slightly worse. For this reason, we compare RAG primarily against ZSI rather than fine-tuned models, since both settings rely on base models without task-specific training.

Overall, these results suggest that the effectiveness of the RAG setup is constrained by the retriever, rather than the reasoning capability of the language model itself.

\begin{table}[H]
\centering
\small
\setlength{\tabcolsep}{3pt}
\renewcommand{\arraystretch}{1.05}
\begin{tabular}{lccc}
\toprule
\textbf{Setup} & \textbf{Precision (m)} & \textbf{Recall (m)} & \textbf{F1 (m)} \\
\midrule
SBERT (Top-15) & 31.99 & 21.23 & 24.36 \\
RAG (LLaMA3)  & 7.51  & 13.00 & 8.46  \\
RAG (Gemma3)  & 5.70  & 6.69  & 5.40  \\
ZSI (LLaMA3)  & 7.70  & 9.40  & 7.04  \\
ZSI (Gemma3)  & 8.98  & 8.32  & 6.94  \\
\bottomrule
\end{tabular}
\caption{SBERT retrieval performance and downstream effects on RAG compared to Zero-Shot Inference (ZSI).}
\label{tab:sbert_rag_analysis}
\end{table}

\begin{table*}[h!]
\centering
\small
\setlength{\tabcolsep}{4pt}
\renewcommand{\arraystretch}{1.05}
\begin{tabular}{l l cc cc cc}
\toprule
\multirow{2}{*}{\textbf{Setup}} & \multirow{2}{*}{\textbf{Model}} 
& \multicolumn{2}{c}{\textbf{Precision}} 
& \multicolumn{2}{c}{\textbf{Recall}} 
& \multicolumn{2}{c}{\textbf{F1}} \\
\cmidrule(lr){3-4} \cmidrule(lr){5-6} \cmidrule(lr){7-8}
& & \textbf{m} & \textbf{$\mu$} & \textbf{m} & \textbf{$\mu$} & \textbf{m} & \textbf{$\mu$} \\
\midrule
Base-RAG & LLaMA3 & 8.56 & 23.06 & 7.83  & 24.84 & 7.30 & 23.92 \\
Base-RAG & Gemma3 & 10.60 & 25.78 & 13.54 & 34.57 & 10.63 & 29.54 \\ 
\midrule
Stat-RAG  & LLaMA3 & 5.63 & 9.24  & 5.34  & 9.33  & 4.08 & 9.29 \\
Stat-RAG  & Gemma3 & 8.94 & 12.49 & 11.40 & 18.05 & 8.29 & 14.76 \\
\midrule
Descr-RAG & LLaMA3 & 7.51 & 13.59 & 13.00 & 31.76 & 8.46 & 19.03 \\
Descr-RAG  & Gemma3 & 5.70 & 13.82 & 6.69  & 17.95 & 5.40 & 15.62 \\
\midrule
Top10-RAG & LLaMA3 & 9.39 & 16.97 & 7.69  & 20.38 & 6.21 & 18.52 \\
Top10-RAG & Gemma3 & 9.39 & 22.20 & 12.53 & 30.87 & 9.57 & 25.83 \\
Top5-RAG  & LLaMA3 & 8.03 & 22.62 & 7.57  & 19.72 & 6.85 & 21.07 \\
Top5-RAG  & Gemma3 & 9.27 & 24.93 & 9.96  & 25.30 & 8.55 & 25.11 \\
\bottomrule
\end{tabular}
\caption{RAG ablation results with macro (m) and micro ($\mu$) evaluation metrics.}
\label{tab:rag_ablation}
\end{table*}

\section{Ablation Study on the RAG Pipeline}
\label{app:rag_ablation}

To better understand the factors affecting the performance of our RAG pipeline, we conduct a series of ablation experiments that systematically modify key components of the retrieval and context construction process. All ablations are evaluated using the same base language models and evaluation protocol as the main RAG setup.
The quantitative results for all ablation settings are reported in Table~\ref{tab:rag_ablation}.

\vspace{3mm}
\noindent \textbf{Ablation on Retrieval Source (Stat-RAG):}
In the RAG setup reported in the main paper, SBERT is used to retrieve relevant evidence by computing similarity between the test query and the set of training queries, followed by selecting the top-15 gold statutes cited by those training queries. As an ablation, we instead retrieve evidence by directly computing SBERT similarity between the test query and the set of all statutes, and select the top-15 statutes to pass to the model. This setting resembles a classical retrieval and re-ranking setup and helps isolate the effect of the retrieval source.

\vspace{3mm}
\noindent \textbf{Ablation on Contextual Information (Descr-RAG):}
In the default RAG setup, we provide title of each retrieved statute as context. To evaluate the impact of contextual richness, we conduct an ablation where both the title and the full statutory description are provided in the prompt.

\vspace{3mm}
\noindent \textbf{Ablation on Evidence Count (Top10-RAG and Top5-RAG): }
The main RAG experiments use the top-15 retrieved statutes as evidence. To study the effect of evidence size, we additionally evaluate RAG setups that use only the top-10 and top-5 retrieved statutes. This ablation helps assess whether reducing the amount of retrieved evidence mitigates noise introduced by imperfect retrieval.

Overall, these ablations allow us to analyze how retrieval strategy, contextual information, and evidence count individually influence the effectiveness of the RAG pipeline.

\section{Qualitative Error Analysis}
\label{app:error_analysis}

To better understand the failure modes of different modeling approaches, we conduct a qualitative error analysis across Zero-Shot Inference (ZSI), Two-Shot Inference (TSI), and Supervised Fine-Tuning (SFT) using the Gemma model. We categorize errors into three broad types based on their relationship to the gold-standard statutes. For each category, we provide a representative example consisting of the input query, the ground-truth statutes, and the model’s prediction.

\vspace{3mm}
\noindent \textbf{Hallucinated Statutes:}
In some cases, the model predicts statutes that do not exist in the Indian legal corpus. These errors indicate hallucination, where the model generates spurious statute references. We observe this error type primarily in ZSI and TSI settings, while SFT does not produce hallucinated statutes in our experiments.  A representative example is provided in \autoref{tab:error_example_zsi}

\begin{table}[h!]
\centering
\small
\setlength{\tabcolsep}{4pt}
\renewcommand{\arraystretch}{1.2}
\begin{tabular}{|p{0.95\columnwidth}|}
\hline
\textbf{Query:} I have agriculture land at village Ottu, Teh. Rania, Distt. Sirsa, Haryana. I own a 4-acre plot that is not connected to any public path. Access is through an approach path owned by my uncle and another person, who often prevent passage. The path has existed for 35 years. Is there any legal or revenue rule that allows me to use this path or obtain legal access to my land, which is 18 karam away and would require crossing my uncle’s land? \\
\\
\textbf{Ground Truth:} Section 15, Section 32, Section 35, and Section 4 of the Indian Easements Act, 1882 \\
\\
\textbf{ZSI Output:} Section 125 and Section 13 of the Hindu Marriage Act, 1955 \\
\hline
\end{tabular}
\caption{Example of hallucinated statute predictions in Zero-Shot Inference (ZSI).}
\label{tab:error_example_zsi}
\end{table}

\noindent The output is a hallucination: the Hindu Marriage Act, 1955 does not contain ``Section 125'', and marriage-related provisions are unrelated to an easement/right-of-way dispute. The correct statutes come from the Indian Easements Act, 1882.

\vspace{3mm}
\noindent \textbf{Non-applicable Statutes:}
In this category, the model predicts statutes that are valid and exist in the legal corpus but are not applicable to the given query and are often unrelated in legal scope. These errors reflect imprecise statute selection, where the model associates a statute with the query despite little or no legal relevance. This error type is observed across ZSI, TSI, and SFT.  A representative example is provided in \autoref{tab:error_example_tsi}.

\begin{table}[h]
\centering
\small
\setlength{\tabcolsep}{4pt}
\renewcommand{\arraystretch}{1.2}
\begin{tabular}{|p{0.95\columnwidth}|}
\hline
\textbf{Query:} SIR, what can I do if a retired army officer is teasing society members, children, and women? We have filed a police case, but no action has been taken as he has strong connections with officials. The person continues to harass children as well. \\
\\
\textbf{Ground Truth:} Section 156 of the Code of Criminal Procedure, 1973 \\
\\
\textbf{TSI Output:} Section 354, Section 355, and Section 356 of the Indian Penal Code, 1860 \\
\hline
\end{tabular}
\caption{Example of non-applicable statute predictions in Two-Shot Inference (TSI).}
\label{tab:error_example_tsi}
\end{table}

\noindent The predicted IPC sections relate to assault and use of criminal force, which are not applicable to the described acts of verbal harassment and intimidation in this query, and therefore do not align with the legal context of the complaint.

\vspace{3mm}
\noindent \textbf{Technically Misaligned Statutes:}
Here, the model predicts statutes that are legally related to the gold standard but differ in technical applicability or scope, reflecting partial understanding of the query. This error type is observed across ZSI, TSI, and SFT.  A representative example is provided in \autoref{tab:error_example_sft}.

\begin{table}[h]
\centering
\small
\setlength{\tabcolsep}{4pt}
\renewcommand{\arraystretch}{1.2}
\begin{tabular}{|p{0.95\columnwidth}|}
\hline
\textbf{Query:} I have agriculture land at village Ottu, Teh. Rania, Distt. Sirsa, Haryana. I own a 4-acre plot that is not connected to any public path. Access is through an approach path owned by my uncle and another person, who often prevent passage. The path has existed for 35 years. Is there any legal or revenue rule that allows me to use this existing path or obtain legal access to my land, which is 18 karam away and would require crossing my uncle’s land? \\
\\
\textbf{Ground Truth:} Section 15, Section 32, Section 35, and Section 4 of the Indian Easements Act, 1882 \\
\\
\textbf{SFT Output:} Section 13 of the Indian Easements Act, 1882; Section 145 of the Code of Criminal Procedure, 1973 \\
\hline
\end{tabular}
\caption{Example of technically misaligned statute predictions in Supervised Fine-Tuning (SFT).}
\label{tab:error_example_sft}
\end{table}

\noindent``Section 13 of The Indian Easements Act, 1882'' is legally related to the dispute but differs in technical applicability from the gold-standard provisions governing prescriptive easements. ``Section 145 of The Code of Criminal Procedure, 1973'' concerns preventive proceedings over breach of peace and is not technically applicable to the underlying right-of-way claim.

Overall, this analysis shows that hallucination is mainly associated with prompting-based methods (ZSI and TSI), while SFT errors primarily arise from non-applicable or partially applicable statute predictions. These findings highlight complementary strengths and weaknesses of prompting and supervised approaches for legal statute identification.










\end{document}